\documentclass{article}

\PassOptionsToPackage{numbers, compress}{natbib}

\usepackage[final]{neurips_2022}

\usepackage{algpseudocode}
\usepackage{subcaption}
\usepackage{algorithm}
\usepackage{mathtools}
\usepackage{graphicx}
\usepackage{makecell}
\usepackage{multirow}
\usepackage{multicol}
\usepackage{caption}
\usepackage{amsmath}
\usepackage{amssymb}
\usepackage{amsthm}
\usepackage{csquotes}
\usepackage{wrapfig}

\usepackage[utf8]{inputenc} 
\usepackage[T1]{fontenc}    
\usepackage{hyperref}       
\usepackage{url}            
\usepackage{booktabs}       
\usepackage{amsfonts}       
\usepackage{nicefrac}       
\usepackage{microtype}      
\usepackage{xcolor}         

\newcommand{\BB}{\mathcal{B}}
\newcommand{\DD}{\mathcal{D}}
\newcommand{\LL}{\mathcal{L}}
\newcommand{\NN}{\mathcal{N}}
\newcommand{\TT}{\mathcal{T}}

\newcommand{\EE}{\mathbb{E}}

\DeclareMathOperator{\clamp}{clamp}

\newtheorem{theorem}{Theorem}
\newtheorem{prop}[theorem]{Proposition}

\title{Energy-Based Contrastive Learning of \\ Visual Representations}

\author{
Beomsu Kim \\
Department of Mathematical Sciences \\
KAIST \\
\texttt{beomsu.kim@kaist.ac.kr} \\
\And
Jong Chul Ye \\
Kim Jaechul Graduate School of AI \\
KAIST \\
\texttt{jong.ye@kaist.ac.kr} \\
}

\begin{document}

\maketitle

\begin{abstract}
Contrastive learning is a method of learning visual representations by training Deep Neural Networks (DNNs) to increase the similarity between representations of positive pairs (transformations of the same image) and reduce the similarity between representations of negative pairs (transformations of different images). Here we explore Energy-Based Contrastive Learning (EBCLR) that leverages the power of generative learning by combining contrastive learning with Energy-Based Models (EBMs). EBCLR can be theoretically interpreted as learning the joint distribution of positive pairs, and it shows promising results on small and medium-scale datasets such as MNIST, Fashion-MNIST, CIFAR-10, and CIFAR-100. Specifically, we find EBCLR demonstrates from $\times 4$ up to $\times 20$ acceleration compared to SimCLR and MoCo v2 in terms of training epochs. Furthermore, in contrast to SimCLR, we observe EBCLR  achieves  nearly the same performance with $254$ negative pairs (batch size $128$) and $30$ negative pairs (batch size $16$) per positive pair, demonstrating the robustness of EBCLR to small numbers of negative pairs. Hence, EBCLR provides a novel avenue for improving contrastive learning methods that usually require large datasets with a significant number of negative pairs per iteration to achieve reasonable performance on downstream tasks. Code: \url{https://github.com/1202kbs/EBCLR}
\end{abstract}

\section{Introduction} \label{sec:intro}

In computer vision, supervised learning requires a large-scale human-annotated dataset of images to train accurate deep neural networks (DNNs). However, acquiring labels for millions of images can be difficult or impossible in practice. This has led to the rise of self-supervised learning, 
which learns useful visual representations by forcing DNNs to be invariant or equivariant to image transformations. Among self-supervised learning algorithms, contrastive methods are rapidly gaining popularity for their superb performance.

Specifically, contrastive learning methods \cite{oord2018,tian2019,tian2020,chen2020,he2020} train DNNs by increasing the similarity between representations of \textit{positive pairs} (transformations of the same image) and decreasing the similarity between representations of \textit{negative pairs} (transformations of different images). The negative pairs prevent DNNs from collapsing to the trivial constant function. There are numerous contrastive learning methods, such as SimCLR \cite{chen2020}, Momentum Contrast (MoCo) \cite{he2020}, etc.

Despite this flurry of research in contrastive learning, contrastive methods require large datasets and a large number of negative pairs per positive pair to achieve reasonable performance on downstream tasks. Although there are recently proposed non-contrastive methods such as BYOL \cite{grill2020} and SimSiam \cite{chen2021} that do not rely on negative pairs, they require heuristic techniques such as stop-gradient to avert collapsing to the trivial solution. There has been an effort to explain the dynamics of non-contrastive methods with linear neural networks \cite{tian2020}, but it is unclear how the analyses generalize to DNNs.
 
\begin{figure}[t]
\centering
\includegraphics[width=0.5\linewidth]{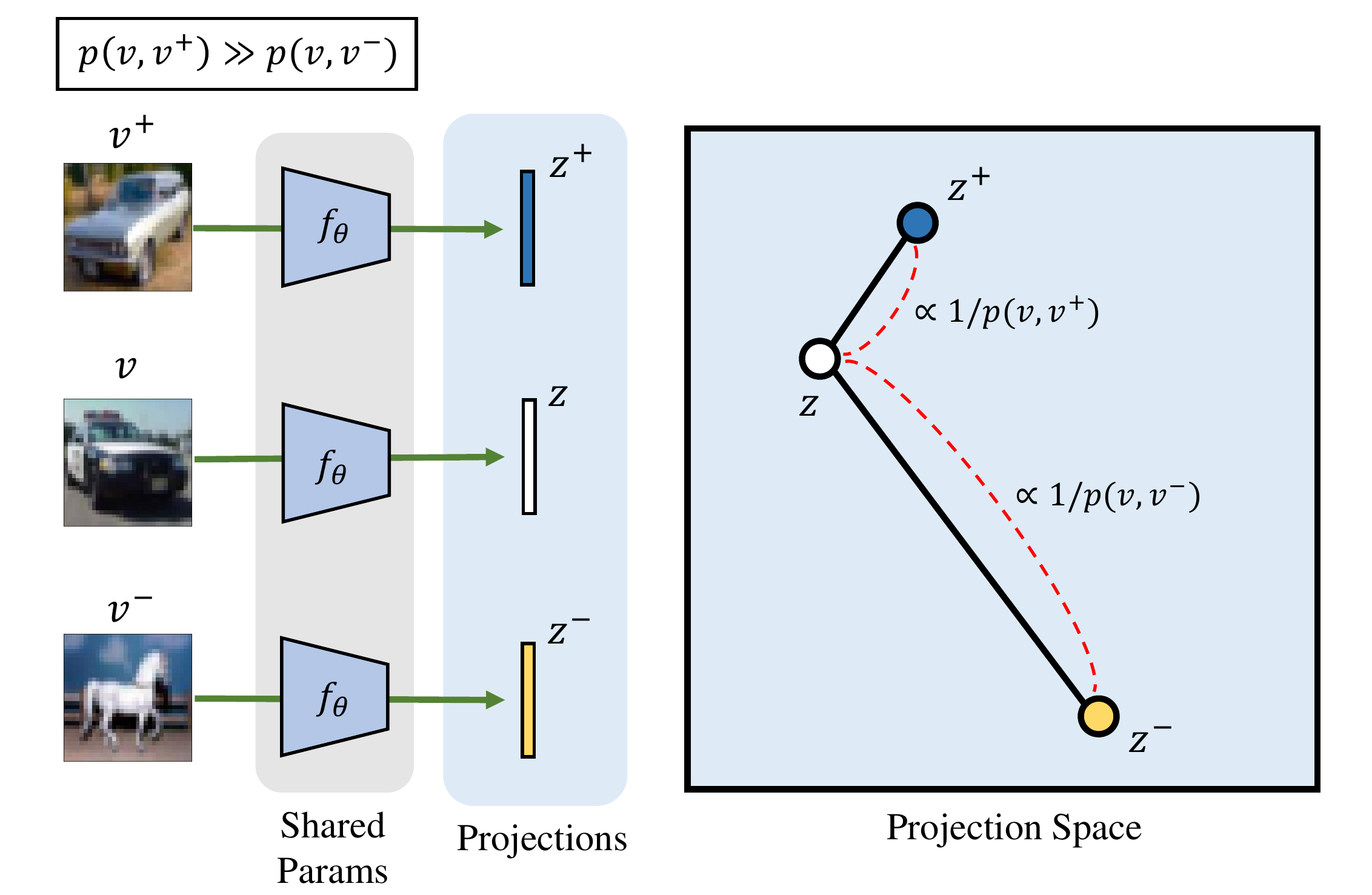} \hfill
\includegraphics[width=0.45\linewidth]{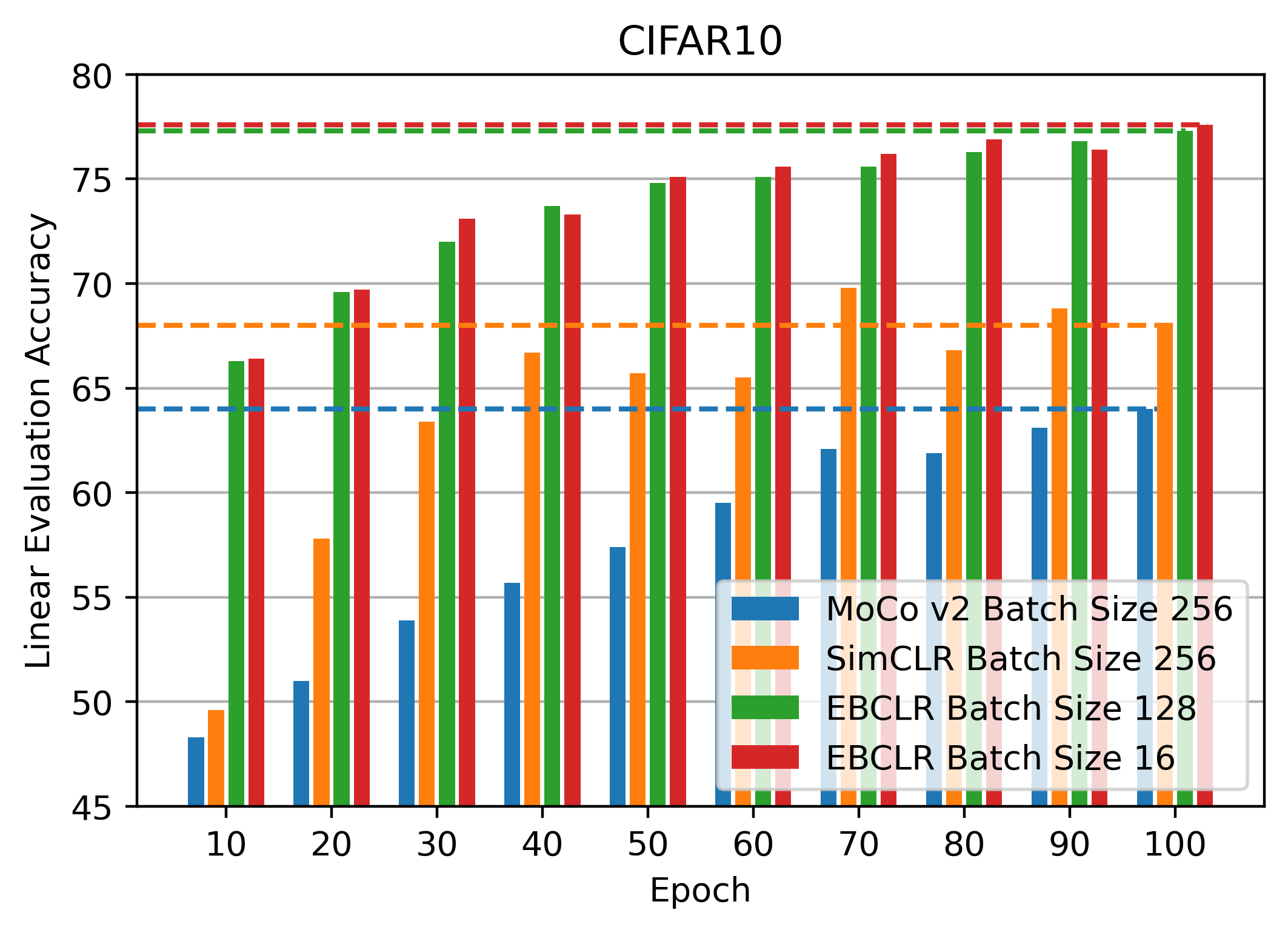}
\caption{\textbf{Left:} An illustration of EBCLR. Here, $\propto$ means ``is a monotonically increasing function of''. We use $p(v,v')$, the joint distribution of positive pairs, as a measure of semantic similarity of images. Specifically, $p(v,v')$ will be high when $v$ and $v'$ are semantically similar, and low otherwise. A DNN $f_\theta$ is trained such that the distance in the projection space is controlled by $1/p(v,v')$. \textbf{Right:} Comparison of EBCLR, SimCLR, and MoCo v2 on CIFAR10 in terms of linear evaluation accuracy. EBCLR at epoch 10 beats MoCo v2 at epoch 100, and EBCLR at epoch 20 beats SimCLR and MoCo v2 at epoch 100. Moreover, EBCLR shows identical performance regardless of whether we use 254 negative pairs (batch size 128) or 30 negative pairs (batch size 16) per positive pair.}
\label{fig:main}
\end{figure}

In this paper, we explore a novel avenue in visual representation learning:  Energy-Based Contrastive Learning (EBCLR) which leverages the power of generative learning \cite{lasserre2006,larochelle2008,grathwohl2020} by combining contrastive learning with energy-based models (EBMs). EBCLR complements the contrastive learning loss with a generative loss, and it can be interpreted as learning the joint distribution of positive pairs. In fact, we demonstrate that the existing contrastive loss is a special
case of the EBCLR loss if the generative term is not used. Although EBMs are notorious for being difficult to train due to their reliance on Stochastic Gradient Langevin Dynamics (SGLD) \cite{welling2011}, 
another important contribution of this work is that
we overcome this by appropriate modifications to SGLD.

Extensive experiments on a variety of small and medium-scale datasets demonstrate that EBCLR is robust to small numbers of negative pairs, and it outperforms SimCLR and MoCo v2 \cite{chen2020moco} in terms of sample efficiency and linear evaluation accuracy. Hence, EBCLR opens up a new research direction for alleviating the dependence of contrastive methods on large datasets and large batches.


Our contributions can be summarized as follows:
\begin{itemize}
\item We propose a novel contrastive learning method called EBCLR which learns the joint distribution of positive pairs. We show that EBCLR loss is equivalent to a combination of a contrastive term and a generative term (Section \ref{sec:theory}). To the best of our knowledge, this is the first work to apply EBMs to contrastive learning of visual representations.
\item We show that EBCLR offers two advantages over conventional contrastive learning methods: EBCLR is several times more sample efficient (Section \ref{sec:exp_comp}) and robust to small batch sizes (Section \ref{sec:exp_neg}). These factors lead to a non-trivial performance gain for EBCLR.
\item We perform thorough ablation studies of the components of EBCLR: effect of changing the weight of the generative term (Section \ref{sec:exp_lmda}), effect of projection space dimension (Section \ref{sec:exp_lmda}), and the effect of the proposed SGLD modifications (Section \ref{sec:exp_msgld}).
\end{itemize}

\section{Related Works} \label{sec:related}

In this section, we go over related works necessary for understanding EBCLR. In Appendix \ref{append:background}, we give a more extensive review of relevant works for those not familiar with EBMs, contrastive learning, or generative models.

\subsection{Contrastive Learning} \label{sec:cl}

For a given batch of images $\{x_n\}_{n = 1}^N$ and two image transformations $t$, $t'$,
contrastive learning methods first create two views $v_n = t(x_n)$, $v_n' = t'(x_n)$ of each instance $x_n$. Here, the pair $(v_n,v_m')$ is called a \textit{positive pair} if $n = m$ and a \textit{negative pair} if $n \neq m$. Given a DNN $f_\theta$, the views are then embedded into the projection space by passing the views through $f_\theta$ and normalizing.

Contrastive methods train $f_\theta$ to increase agreement between projections of positive pairs and decrease agreement between projections of negative pairs. Specifically, $f_\theta$ is trained to maximize the InfoNCE objective \cite{oord2018}. After training, outputs from the final layer or an intermediate layer of $f_\theta$ are used for downstream tasks.

There are numerous variants of contrastive methods. For instance, SimCLR \cite{chen2020} uses a composition of random cropping, random flipping, color jittering, color dropping, and blurring as the image transformation. Negative pairs are created by transforming different images within a batch. On the other hand, MoCo \cite{chen2020moco} maintains a queue of negative samples, so negative samples are not limited to views of images from the same batch.

\subsection{Energy-Based Models} \label{sec:ebm}

Given a scalar-valued \textit{energy function} $E_\theta(v)$ with parameter $\theta$, an energy-based model (EBM) \cite{lecun2006} defines a distribution by the formula
\begin{align}
q_\theta(v) \coloneqq \frac{1}{Z(\theta)} \exp\{-E_\theta(v)\} \label{eq:ebm}
\end{align}
where $Z(\theta)$ is the \textit{partition function} which guarantees $q_\theta$ integrates to $1$. Since there are essentially no restrictions on the choice of the energy function, EBMs have great flexibility in modeling distributions. Hence, EBMs have been applied to a wide variety of machine learning tasks, such as dimensionality reduction via autoencoding \cite{ranzato2007}, learning generative classifiers \cite{lasserre2006,larochelle2008,grathwohl2020,yang2021}, generating images \cite{du2019}, and training regression models \cite{gustafsson2020a,gustafsson2020b}. Wang et al. \citep{wang2022} have explored connections between EBMs and InfoNCE to enhance generative performance of EBMs. However, to the best of our knowledge, this paper is the first to combine EBMs with contrastive learning for representation learning.

Given a target distribution, an EBM can be used to estimate its density $p$ when we can only sample from $p$. One way of achieving this is
by minimizing the Kullback-Leibler (KL) divergence between $q_\theta$ and $p$ that maximizes the expected log-likelihood of $q_\theta$ under $p$ \cite{song2021}:
\begin{align}
\max_\theta \ \EE_p[\log q_\theta(v)]. \label{eq:ell}
\end{align}

Stochastic gradient ascent can be used to solve \eqref{eq:ell} \cite{song2021}. 
Specifically, the gradient of the expected log-likelihood with respect to the parameters $\theta$ can be shown to be
\begin{align}
\nabla_\theta \EE_p[\log q_\theta(v)] = \EE_{q_\theta}[\nabla_\theta E_\theta(v)] - \EE_{p}[\nabla_\theta E_\theta(v)]. \label{eq:ebm_grad}
\end{align}
Hence, updating $\theta$ with \eqref{eq:ebm_grad} amounts to pushing up on the energy for samples from $q_\theta$ and pushing down on the energy for samples from $p$. This optimization method is also known as contrastive divergence \cite{hinton2002}.

While the second term in \eqref{eq:ebm_grad} can be easily calculated as we have access to samples from $p$, the first term requires sampling from $q_\theta$. Previous works \cite{du2019,nijkamp2019,grathwohl2020,yang2021} have used Stochastic Gradient Langevin Dynamics (SGLD) \cite{welling2011} to generate samples from $q_\theta$. Specifically, given a sample $v_0$ from some proposal distribution $q_0$,  the iteration
\begin{align}
v_{t + 1} = v_t - \frac{\alpha_t}{2} \nabla_{v_t} E_\theta(v_t) + \epsilon_t, \quad \epsilon_t \sim \NN(0,\sigma_t^2) \label{eq:sgld}
\end{align}
guarantees that the sequence $\{v_t\}$ converges to a sample from $q_\theta$ assuming $\{\alpha_t\}$ decays at a polynomial rate \cite{welling2011}.

However, SGLD requires an infinite number of steps until samples from the proposal distribution converge to samples from the target distribution. This is unfeasible, so in practice, only a finite number of steps along with constant step size, i.e. $\alpha_t=\alpha$ and constant noise variance $\sigma_t=\sigma^2$ are used \cite{du2019,nijkamp2019,grathwohl2020,yang2021}.
Moreover, Yang and Ji \cite{yang2021} noted SGLD often generates samples with extreme pixel values that cause EBMs to diverge during training. Hence, they have proposed \textit{proximal SGLD} which clamps gradient values into an interval $[-\delta,\delta]$ for a threshold $\delta > 0$. Then, the update equation becomes
\begin{align}
v_{t + 1} = v_t - \alpha \cdot \clamp\{\nabla_v E_\theta(v_t),\delta\} + \epsilon \label{eq:prox_sgld}
\end{align}
for $t = 0, \ldots, T - 1$,
where $\epsilon \sim \NN(0, \sigma^2)$ and $\clamp\{\cdot,\delta\}$ clamps each element of the input vector into $[-\delta,\delta]$. In our work, we introduce additional modifications to SGLD which accelerate the convergence of EBCLR.

\begin{figure*}[t!]
\centering
\includegraphics[width=1.0\linewidth]{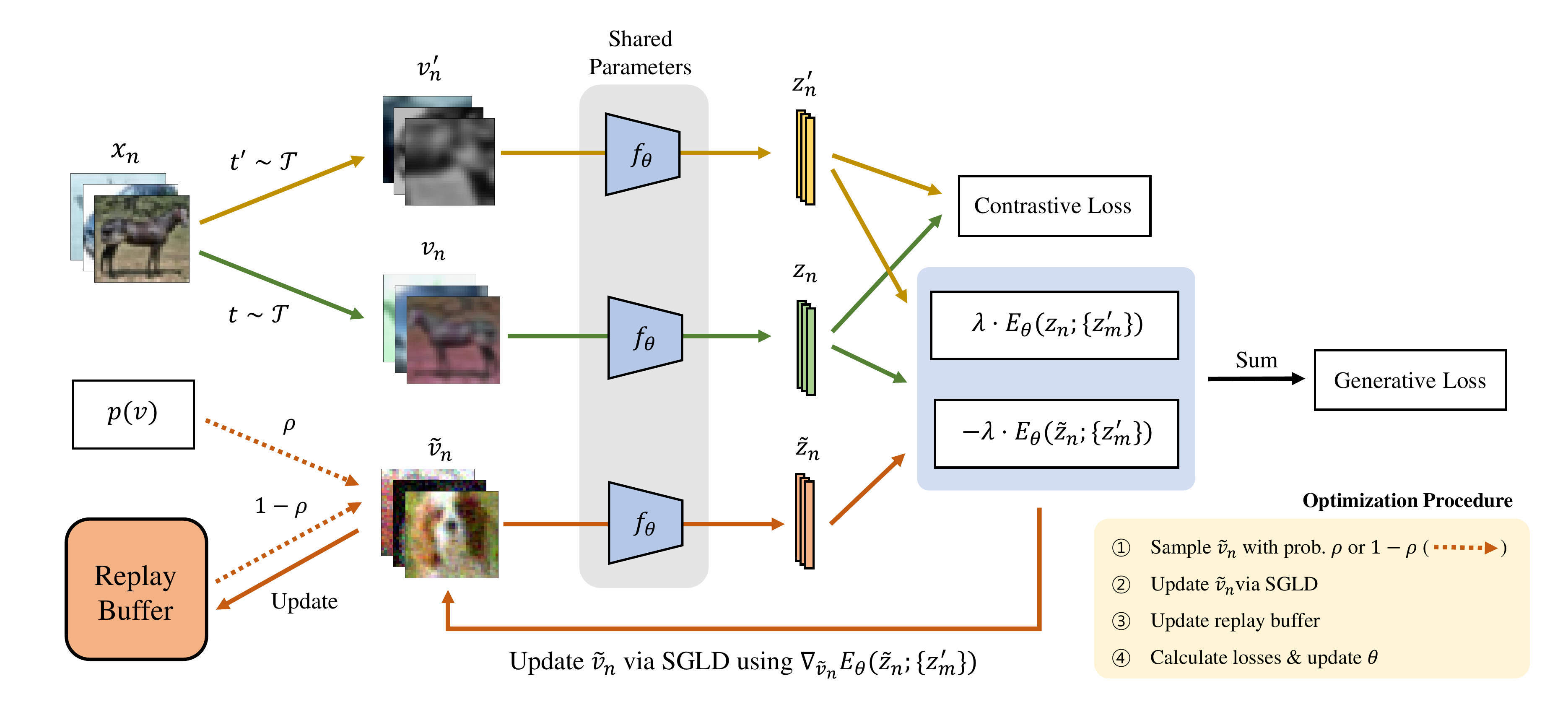}
\caption{An illustration of the learning process of EBCLR.}
\label{fig:ebclr_flow}
\end{figure*}

\section{Theory} \label{sec:theory}

\subsection{Energy-Based Contrastive Learning}

Let $\DD$ be a distribution of images and $\TT$ a distribution of stochastic image transformations. Given $x \sim \DD$ and i.i.d. $t,t'\sim \TT$, our goal is to approximate the joint distribution of the views
$$p(v,v'), \quad \mbox{where}~~v=t(x),~v'=t'(x)$$
using the model distribution
\begin{align}
q_\theta(v,v') \coloneqq \frac{1}{Z(\theta)} \exp\{-\| z - z' \|^2 / \tau\}. \label{eq:joint}
\end{align}
where $Z(\theta)$ is a normalization constant, $\tau > 0$ is a temperature hyper-parameter, and $z$ and $z'$ are projections computed by passing the views $v$ and $v'$ through the DNN $f_\theta$ and then normalizing to have unit norm. We now explain the intuitive meaning of matching $q_\theta$ to $p$.

Our key idea is to use $p(v,v')$ as a measure of semantic similarity of $v$ and $v'$. If two images $v$ and $v'$ are semantically similar, they are likely to be transformations of similar images. So, $p(v,v')$ will be high when $v$ and $v'$ are semantically similar and low otherwise. Suppose $q_\theta$ successfully approximates $p$. If we equate $p(v,v')$ to $q_\theta(v,v')$ in \eqref{eq:joint} and solve for $\|z - z'\|$, we see that the distance between $z$ and $z'$ will become a monotone increasing function of $1/p(v,v')$, which is the inverse of semantic similarity of $v$ and $v'$. So, semantically similar images will have nearby projections, and dissimilar images will have distant projections. This idea is illustrated in Figure \ref{fig:main}.

To approximate $p$ using $q_\theta$, we train $f_\theta$ 
to maximize the expected log-likelihood of $q_\theta$ under $p$:
\begin{align}\label{eq:original}
\max_\theta \ \EE_{p}[\log q_\theta(v,v')].
\end{align}
In order to solve this problem with stochastic gradient ascent, we could naively extend \eqref{eq:ebm_grad} to the setting of joint distributions to obtain the following result.
\begin{prop}
The the joint distribution \eqref{eq:joint} can be formulated as an EBM
\begin{align}
q_\theta(v,v') = \frac{1}{Z(\theta)} \exp\{-E_\theta(z,z')\}, \qquad E_\theta(v,v') = \|z - z'\|^2 / \tau
\label{eq:joint_energy}
\end{align}
and the gradient of the objective of \eqref{eq:original} is given by
\begin{align}
\nabla_\theta \EE_{p}[\log q_\theta(v,v')] = \EE_{q_\theta}[\nabla_\theta E_\theta(v,v')] - \EE_{p}[\nabla_\theta E_\theta(v,v')].
\label{eq:joint_grad}
\end{align}
\end{prop}

However, computing the first expectation in \eqref{eq:joint_grad} requires sampling pairs of views $(v,v')$ from $q_\theta(v,v')$ via SGLD, which could be expensive. To avert this problem, we use Bayes' rule to decompose
\begin{align}
\EE_{p}[\log q_\theta(v,v')] = \EE_p [\log q_\theta(v' \mid v)] + \EE_p [\log q_\theta(v)] \quad \text{where} \quad q_\theta(v) = \int q_\theta(v,v') \, dv'. \label{eq:decomp}
\end{align}
In the first equation of \eqref{eq:decomp}, the first and second terms at the RHS will be referred to as discriminative and generative terms, respectively, throughout the paper.
A similar decomposition was used by Grathwohl et al. \cite{grathwohl2020} in the setting of learning generative classifiers.

Furthermore, we add a hyper-parameter $\lambda$ to balance the strength of the discriminative term and the generative term. The advantage of this modification will be discussed in Section \ref{sec:exp_lmda}. This yields our Energy-Based Contrastive Learning (EBCLR) objective
\begin{align}
\LL(\theta) \coloneqq \EE_p [\log q_\theta(v' \mid v)] + \lambda \EE_p [\log q_\theta(v)]. \label{eq:ebclr}
\end{align}

The discriminative term can be easily differentiated since the partition function $Z(\theta)$ cancels out when $q_\theta(v,v')$ is divided by $q_\theta(v)$. However, the generative term still contains $Z(\theta)$. We now present our key result, which is used to maximize \eqref{eq:ebclr}. The proof is deferred to Appendix \ref{append:thm2_proof}.

\begin{theorem} \label{thm:2}
The marginal distribution in \eqref{eq:decomp} can be formulated as an EBM
\begin{align}
q_\theta(v) = \frac{1}{Z(\theta)} \exp\{-E_\theta(v)\}, \qquad E_\theta(v) \coloneqq - \log \int e^{-\|z- z'\|^2 / \tau} \, dv' \label{eq:marginal_ebm}
\end{align}
where $Z(\theta)$ is the partition function in \eqref{eq:joint}, and the gradient of the generative term is given by
\begin{align}
\nabla_\theta \EE_p[\log q_\theta(v)] = \EE_{q_\theta(v)}[\nabla_\theta E_\theta(v)] - \EE_p [\nabla_\theta E_\theta(v)]. \label{eq:marginal_grad}
\end{align}
Thus, the gradient of the EBCLR objective is
\begin{align}
\nabla_\theta \LL(\theta) = \EE_p [\nabla_\theta \log q_\theta(v' \mid v)] + \lambda \EE_{q_\theta(v)}[\nabla_\theta E_\theta(v)] - \lambda \EE_{p}[\nabla_\theta E_\theta(v)] \label{eq:ebclr_grad}
\end{align}
\end{theorem}

Theorem~\ref{thm:2} suggests that the EBM for the joint distribution can be learned by computing the gradients of the discriminative term and the EBM for the marginal distribution. Moreover, we only need to sample $v$ from $q_\theta(v)$ to compute the second expectation in \eqref{eq:ebclr_grad}.

\subsection{Approximating the EBCLR Objective}

To implement EBCLR, we need to approximate expectations in \eqref{eq:ebclr} with their empirical means. Suppose samples $\{(v_n,v'_n)\}_{n = 1}^N$ from $p(v,v')$ are given,
and  let $\{(z_n,z'_n)\}_{n = 1}^N$ be the corresponding projections.
As the learning goal is to make $q_\theta(v_n, v'_n)$ approximate the joint probability density function $p(v_n, v'_n)$,
the empirical mean $\widehat q_\theta(v_n)$ can be defined as:
\begin{align}
\widehat q_\theta(v_n) =  & \frac{1}{N'}  
\sum_{v_m' : v'_m \neq v_n} q_\theta(v_n, v'_m) \label{eq:approx2} 
\end{align}
where the sum is over the collection of $v_m'$ defined as
\begin{align}
\{v_m' : v'_m \neq v_n\} \coloneqq \{v_k\}_{k = 1}^N \cup \{v_k'\}_{k = 1}^N - \{v_n\}
\end{align}
and $N' \coloneqq |\{v_m' : v'_m\neq v_n\}| = 2N-1$.
One could also use a simpler form of the 
empirical mean:
\begin{align}
\widehat q_\theta(v_n) =  & \frac{1}{N}  
\sum_{m = 1}^N q_\theta(v_n, v'_m) \label{eq:approx1} 
\end{align}
Similarly, 
$q_\theta(v' | v)$ in \eqref{eq:ebclr}, which should approximate the conditional probability density $p(v' | v)$, can be represented in terms of $q_\theta(v_n, v'_n)$. Specifically, we have
\begin{align}
q_\theta(v_n' \mid v_n) \simeq \frac{q_\theta(v_n, v'_n)}{\widehat q_\theta(v_n)} &=\frac{q_\theta(v_n,v'_n)}{\frac{1}{N'} \sum_{v_m' : v'_m \neq v_n} q_\theta(v_n,v'_m)} = \frac{e^{-\| z_n - z'_n \|^2 / \tau}}{\frac{1}{N'} \sum_{v_m' : v'_m\neq v_n} e^{-\| z_n - z'_m \|^2 / \tau}}
\label{eq:q_bar_approx}
\end{align}

It is then immediately apparent that the empirical form of the discriminative term using \eqref{eq:q_bar_approx} is a particular instance of the contrastive learning objective such as InfoNCE and SimCLR. Hence, EBCLR can be interpreted as complementing contrastive learning with a generative term defined by an EBM. We will demonstrate in Section \ref{sec:exp_comp} that the generative term offers significant advantages over other contrastive learning methods.

For the second term, we use the simpler form of the empirical mean in \eqref{eq:approx1}:
\begin{align}
\widehat q_\theta(v_n) = \frac{1}{N} \sum_{m = 1}^N q_\theta(v_n,v'_m) = \frac{1}{Z(\theta)} \cdot \frac{1}{N} \sum_{m = 1}^N \exp\{- \| z_n - z'_m \|^2 / \tau\}
\label{eq:q_approx}
\end{align}
We could also use \eqref{eq:approx2} as the empirical mean, but either choice showed identical performance (see Appendix \ref{sec:approx_comp}). So, we have found \eqref{eq:approx2} to be not worth the additional complexity, and have resorted to the simpler approximation \eqref{eq:approx1} instead. In Appendix \ref{append:another}, we theoretically justify that EBCLR will work as intended even with the approximations \eqref{eq:approx2} or \eqref{eq:approx1}. If we compare \eqref{eq:q_approx} with \eqref{eq:marginal_ebm}, we can see that this approximation of $q_\theta(v)$ yields the  energy function (after ignoring the constant $\log N$)
\begin{align}
E_\theta(v;\{v'_m\}_{m = 1}^N) \coloneqq - \log \left(\sum_{m = 1}^N e^{- \| z - z'_m \|^2 / \tau }\right). \label{eq:energy_approx}
\end{align}

\subsection{Modifications to SGLD} \label{sec:sgld_mod}

According to Theorem~\ref{thm:2}, we need samples from the marginal 
$q_\theta(v)$ to calculate the second expectation in \eqref{eq:ebclr_grad}. Hence, we apply proximal SGLD \eqref{eq:prox_sgld} with the energy function \eqref{eq:energy_approx} to sample from $q_\theta(v)$ as
\begin{align}
\tilde{v}_{t + 1} = \tilde{v}_t - \alpha \cdot \clamp\{\nabla_{v} E_\theta(\tilde{v}_t;\{v'_m\}_{m = 1}^N),\delta\} + \epsilon
\end{align}
for $t = 0, \ldots, T - 1$, where $\epsilon \sim \NN(0,\sigma^2)$. We make three additional modifications to proximal SGLD to expedite the training process. From here on, we will be referring to proximal SGLD in \eqref{eq:prox_sgld} when we say SGLD.

First, we initialize SGLD from generated samples from previous iterations, and with probability $\rho$, we reinitialize SGLD chains from samples from a proposal distribution $q_0$. This is achieved by keeping a replay buffer $\BB$ of SGLD samples from previous iterations. This technique of maintaining a replay buffer has also been used in previous works and has proven to be crucial for stabilizing and accelerating the convergence of EBMs \cite{du2019,grathwohl2020,yang2021}.

Second, the proposal distribution $q_0$ is set to be the data distribution $p(v)$. This choice differs from those of previous works \cite{du2019,grathwohl2020,yang2021} which have either used the uniform distribution or a mixture of Gaussians as the proposal distribution.

Finally, we use \textit{multi-stage SGLD (MSGLD)}, which adaptively controls the magnitude of noise added in SGLD. For each sample $\tilde{v}$ in the replay buffer $\BB$, we keep a count $\kappa_{\tilde{v}}$ of number of times it has been used as the initial point of SGLD. For samples with a low count, we use noise of high variance, and for samples with a high count, we use noise of low variance. Specifically, in \eqref{eq:prox_sgld}, we set
\begin{align}
\sigma = \sigma_{\min} + (\sigma_{\max} - \sigma_{\min}) \cdot [1 - \kappa_{\tilde{v}} / K]_+.
\end{align}
where $[\cdot]_+ \coloneqq \max\{0,\cdot\}$,  $\sigma_{\max}^2$ and $\sigma_{\min}^2$ are the upper and lower bounds on the noise variance, respectively, and $K$ controls the decay rate of noise variance. The purpose of this technique is to facilitate quick exploration of the modes of $q_\theta$ and still guarantee SGLD generates samples with sufficiently low energy. The pseudocodes for MSGLD and EBCLR are given in Algorithms \ref{alg:msgld} and \ref{alg:ebclr}, respectively, in Appendix \ref{append:psuedocodes}, and the overall learning flow of EBCLR is described in Figure~\ref{fig:ebclr_flow}.

\section{Experiments} \label{sec:exp}

We now describe the experimental settings. A complete description is deferred to Appendix \ref{append:details}.

\textbf{Baseline methods and datasets.} The baseline methods are SimCLR, MoCo v2, SimSiam, and BYOL. The hyper-parameters are chosen closely following the original works \cite{chen2020,chen2020moco,chen2021,grill2020}. We use four datasets: MNIST \cite{mnist}, Fashion MNIST (FMNIST) \cite{fmnist}, CIFAR10, and CIFAR100 \cite{cifar}.

\textbf{DNN architecture.} We decompose $f_\theta = \pi_\theta \circ \phi_\theta$ where $\phi_\theta$ is the encoder network and $\pi_\theta$ is the projection network. Rather than using the output of $f_\theta$ for downstream tasks, we follow previous works \cite{chen2020,he2020,oord2018,tian2019,tian2020,grill2020,chen2021} and use the output of $\phi_\theta$ instead. In our experiments, we set $\phi_\theta$ to be a ResNet-18 \cite{he2016} up to the global average pooling layer and $\pi_\theta$ to be a 2-layer MLP with output dimension $128$. However, we remove batch normalization because batch normalization hurts SGLD \cite{du2019}. We also replace ReLU with leaky ReLU to expedite the convergence of SGLD. For the baselines, we use settings proposed in the original works while keeping the backbone fixed to be ResNet-18.

\textbf{Evaluation.} We evaluate the representations by training a linear classifier on top of frozen $\phi_\theta$.

\begin{table*}[t!]
\centering
\resizebox{0.95\textwidth}{!}{
\begin{tabular}{c c c c c c c c c}
\toprule
Dataset & \multicolumn{2}{c}{\textbf{MNIST}} & \multicolumn{2}{c}{\textbf{FMNIST}} & \multicolumn{2}{c}{\textbf{CIFAR10}} & \multicolumn{2}{c}{\textbf{CIFAR100}} \\
\cmidrule(lr){1-1} \cmidrule(lr){2-3} \cmidrule(lr){4-5} \cmidrule(lr){6-7} \cmidrule(lr){8-9}
Statistic & Accuracy & Rel. Eff. & Accuracy & Rel. Eff. & Accuracy & Rel. Eff. & Accuracy & Rel. Eff. \\
\cmidrule{1-9}
SimSiam & $98.6$ & $0.1$ & $87.4$ & $0.1$ & $70.4$ & $0.25$ & $38.3$ & $0.1$ \\
BYOL & $\mathbf{99.3}$ & $0.4$ & $89.0$ & $0.2$ & $70.9$ & $0.25$ & $41.7$ & $0.2$ \\
SimCLR  & $99.0$ & $0.1$ & $88.5$ & $0.15$ & $68.0$ & $0.15$ & $43.1$ & $0.25$ \\
MoCo v2 & $98.1$ & $0.05$ & $87.8$ & $0.1$ & $64.0$ & $0.1$ & $38.2$ & $0.1$ \\
\cmidrule{1-9}
EBCLR   & $\mathbf{99.3}$ & -- & $\mathbf{90.1}$ & -- & $\mathbf{77.3}$ & -- & $\mathbf{49.1}$ & -- \\
\bottomrule
\end{tabular}}
\caption{Linear evaluation accuracy and efficiency relative to EBCLR. Efficiency of a method relative to EBCLR is calculated by the following formula: (number of epochs used by EBCLR to reach the final accuracy of the method) / (total number of training epochs).}
\label{table:baseline}
\end{table*}

\begin{wraptable}{r}{0.55\linewidth}
\centering
\resizebox{0.5\textwidth}{!}{
\begin{tabular}{c c c c c}
\toprule
Direction & \textbf{M} $\rightarrow$ \textbf{FM} & \textbf{FM} $\rightarrow$ \textbf{M} & \textbf{C10} $\rightarrow$ \textbf{C100} & \textbf{C100} $\rightarrow$ \textbf{C10} \\
\cmidrule{1-5}
SimSiam & $86.9$ & $97.2$ & $39.5$ & $64.0$ \\
BYOL    & $\mathbf{87.3}$ & $97.8$ & $42.3$ & $70.2$ \\
SimCLR  & $86.9$ & $97.4$ & $39.9$ & $67.3$ \\
MoCo v2 & $85.3$ & $97.1$ & $36.2$ & $62.9$ \\
\cmidrule{1-5}
EBCLR   & $\mathbf{87.4}$ & $\mathbf{98.5}$ & $\mathbf{46.9}$ & $\mathbf{72.4}$ \\
\bottomrule
\end{tabular}}
\vspace{1em}
\caption{Comparison of transfer learning results in the linear evaluation setting. Left side of the arrow is the dataset than the encoder was pre-trained on, and right side of the arrow is the dataset that linear evaluation was performed on. We use the following abbreviations. \textbf{M} : MNIST, \textbf{FM} : FMNIST, \textbf{C10} : CIFAR10, \textbf{C100} : CIFAR100.}
\label{table:transfer}
\end{wraptable}

\subsection{Comparison with Baselines} \label{sec:exp_comp}

We use batch size 128 for EBCLR and batch size 256 for the baseline methods following Wang et. al \cite{wang2021} and train each method for 100 epochs. Table \ref{table:baseline} shows the result of training each method for 100 epochs. Observe that EBCLR consistently outperforms all baseline methods in terms of linear evaluation accuracy. Moreover, relative efficiency indicates EBCLR is capable of achieving the same level of performance as the baseline methods with much fewer training epochs. Concretely, we observe at least $\times 4$ acceleration in terms of epochs compared to contrastive methods. Hence, EBCLR is a much more desirable choice than SimCLR or MoCo v2 for learning visual representations when we have a small number of training samples.
 
We also investigate the transfer learning performance of EBCLR. Table \ref{table:transfer} compares the transfer learning accuracies. EBCLR always outperforms the baseline methods, and the performance gap is especially large on CIFAR10 and CIFAR100. This indicates EBCLR learns visual representations that generalize well across datasets. Repeating the above experiments with longer training or KNN classification led to similar conclusions (see Appendixes \ref{append:exp_comp} and \ref{append:knn}, respectively).

\subsection{Effect of Reducing Negative Pairs} \label{sec:exp_neg}

We compared the performances of EBCLR and SimCLR as we reduced the number of negative pairs per positive pair. For MoCo v2, the negative samples are provided by a queue updated by a momentum encoder. On the other hand, for EBCLR and SimCLR, negative samples come from the same batch as the positive pair. So, we did not have a way of fairly comparing EBCLR and SimCLR with MoCo v2. Hence, we excluded MoCo v2 from this experiment.

We note that, according to \eqref{eq:q_bar_approx}, given a batch of size $N$, we obtain $2N - 2$ negative pairs for each positive pair. SimCLR also has $2N - 2$ negative pairs for each positive pair. Hence, we can conveniently compare the sensitivity of EBCLR and SimCLR to the number of negative pairs by varying the batch size.

Table \ref{table:batch} shows the result of training each method for $100$ epochs with batch sizes in $\{16, 64, 128\}$. We make three important observations. First, EBCLR consistently beats SimCLR in terms of linear evaluation accuracy for every batch size. Second, EBCLR is invariant to the choice of batch size. This contrasts with SimCLR whose performance degrades as batch size decreases. Consequently, EBCLR with batch size $16$ beats SimCLR with batch size $128$. Finally, as a byproduct of the second observation, the efficiency of EBCLR relative to SimCLR increases as batch size decreases. These properties make EBCLR suitable for situations where we cannot use large batch sizes, e.g., when we have a small number of GPUs. Repeating the experiments with longer training or KNN classification again led to similar conclusions (see Appendixes \ref{append:exp_neg} and \ref{append:knn}, respectively).

\begin{table*}[t!]
\centering
\resizebox{0.95\textwidth}{!}{
\begin{tabular}{c c c c c c c c c c c c c}
\toprule
Dataset & \multicolumn{3}{c}{\textbf{MNIST}} & \multicolumn{3}{c}{\textbf{FMNIST}} & \multicolumn{3}{c}{\textbf{CIFAR10}} & \multicolumn{3}{c}{\textbf{CIFAR100}} \\
\cmidrule(lr){1-1} \cmidrule(lr){2-4} \cmidrule(lr){5-7} \cmidrule(lr){8-10} \cmidrule(lr){11-13}
Batch Size & $16$ & $64$ & $128$ & $16$ & $64$ & $128$ & $16$ & $64$ & $128$ & $16$ & $64$ & $128$ \\
\cmidrule{1-13}
SimCLR & $98.7$ & $99.1$ & $99.1$ & $87.1$ & $88.0$ & $88.2$ & $65.2$ & $67.6$ & $69.0$ & $36.9$ & $39.1$ & $43.0$ \\
EBCLR & $\mathbf{99.4}$ & $\mathbf{99.3}$ & $\mathbf{99.3}$ & $\mathbf{89.6}$ & $\mathbf{90.4}$ & $\mathbf{90.1}$ & $\mathbf{77.6}$ & $\mathbf{78.2}$ & $\mathbf{77.3}$ & $\mathbf{48.8}$ & $\mathbf{49.8}$ & $\mathbf{49.1}$ \\
\cmidrule{1-13}
Rel. Eff. & $0.05$ & $0.1$ & $0.15$ & $0.05$ & $0.15$ & $0.1$ & $0.1$ & $0.15$ & $0.2$ & $0.1$ & $0.15$ & $0.25$ \\
\bottomrule
\end{tabular}}
\caption{Linear evaluation accuracies and efficiencies relative to EBCLR with various batch sizes. Efficiency of SimCLR relative to EBCLR is calculated by the following formula: (number of epochs used by EBCLR with the same batch size to reach the final accuracy of SimCLR) / (total number of training epochs).}
\label{table:batch}
\end{table*}

\begin{wrapfigure}{r}{0.5\linewidth}
\centering
\begin{subfigure}{0.49\linewidth}
\includegraphics[width=1.0\linewidth]{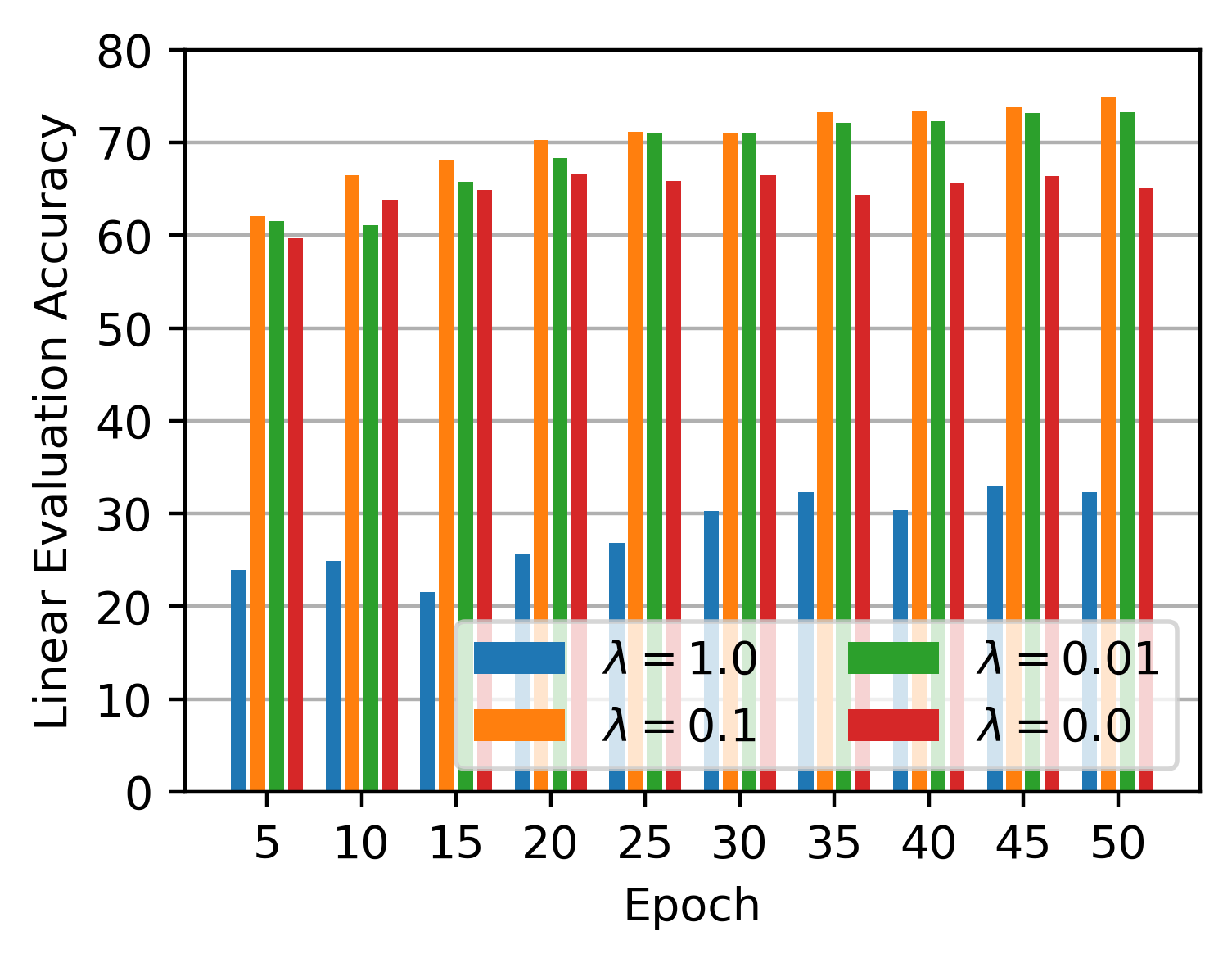}
\caption{Effect of $\lambda$.}
\label{fig:lmda}
\end{subfigure}
\begin{subfigure}{0.49\linewidth}
\includegraphics[width=1.0\linewidth]{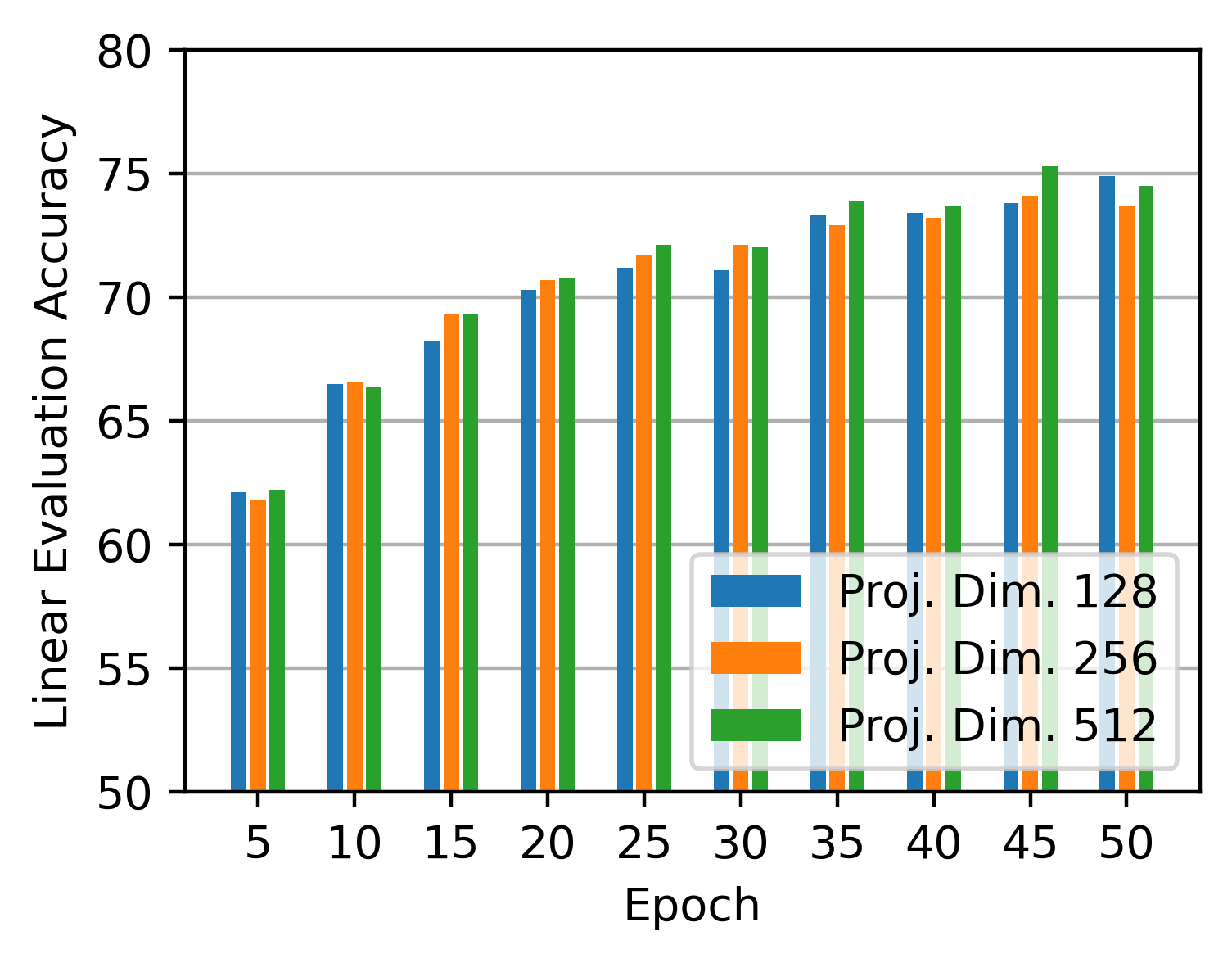}
\caption{Effect of proj. dim.}
\label{fig:proj}
\end{subfigure}
\caption{Effect of $\lambda$ and projection dimension (output dimension of $\pi_\theta$), demonstrated on CIFAR10.}
\end{wrapfigure}

\subsection{Effect of $\lambda$ and Projection Dimension} \label{sec:exp_lmda}

We explored the effect of changing the hyper-parameter $\lambda$ which controls the importance of the generative term relative to the discriminative term (see Equation \eqref{eq:ebclr}). Figure \ref{fig:lmda} shows the performance of EBCLR with various values of $\lambda$ as training progresses. We observe that naively using $\lambda = 1.0$ leads to poor results. The performance peaks at $\lambda = 0.1$, and then degrades as we further decrease $\lambda$.

This result has two crucial implications. First, the generative term plays a non-trivial role in EBCLR. Second, we need to strike a right balance between the discriminative term and the generative term to achieve good performance on downstream tasks\footnote{Interestingly, we observed a similar phenomenon when we used models trained with EBCLR to generate images. For more details, we refer the readers to Appendix \ref{append:exp_gen}.}.

We also investigated the effect of varying the output dimension of $\pi_\theta$. Figure \ref{fig:proj} shows linear evaluation results for projection dimensions in $\{128, 256, 512\}$. We observe that the projection dimension has essentially no influence on the training process. In this respect, EBCLR resembles SimCLR which is also invariant to the output dimension (see Figure 8 in the work by Chen et. al \cite{chen2020}).

\subsection{Effect of SGLD Modifications} \label{sec:exp_msgld}

\begin{wrapfigure}{r}{0.45\linewidth}
\centering
\begin{subfigure}{0.49\linewidth}
\includegraphics[width=1.0\linewidth]{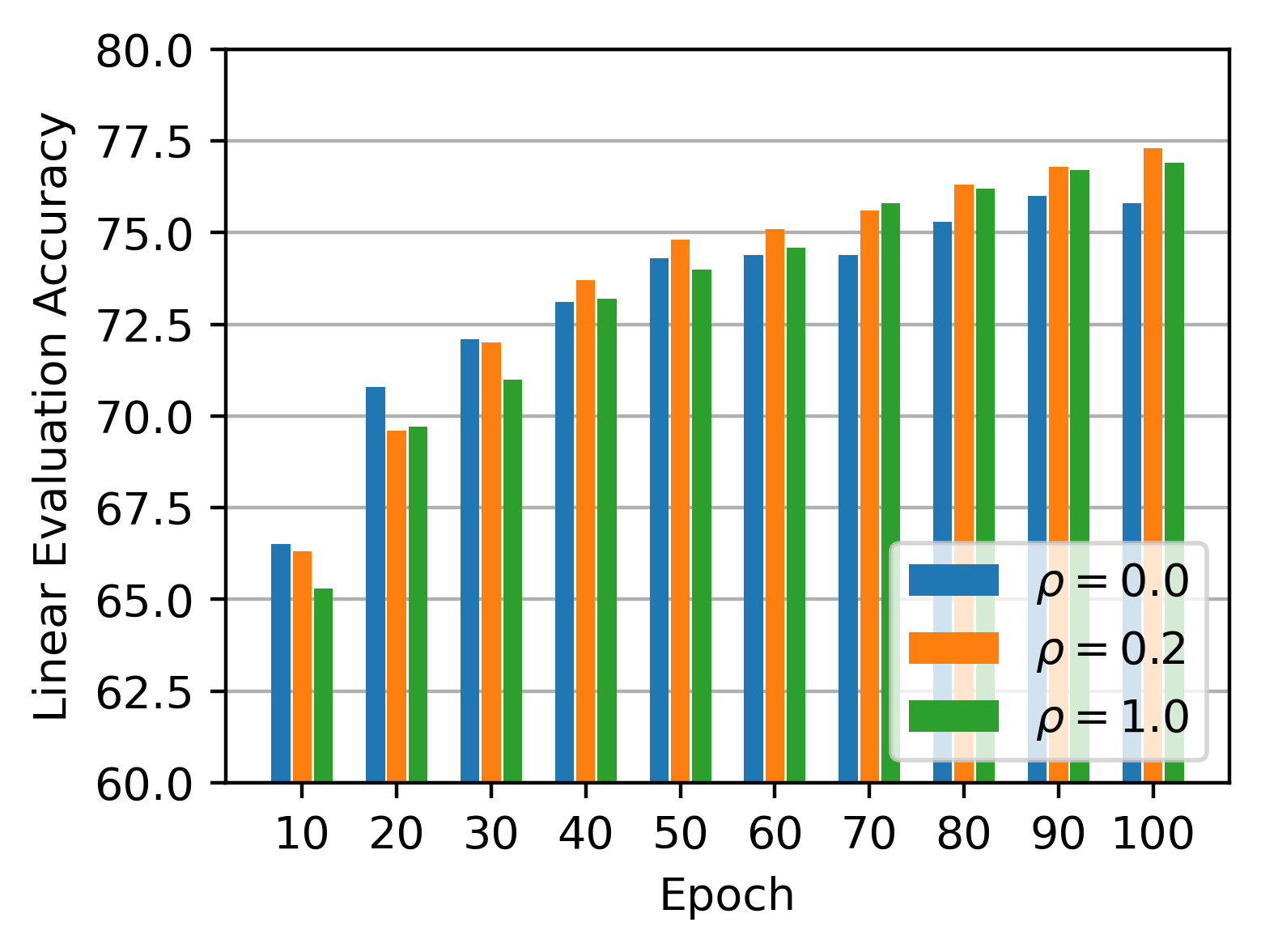}
\caption{Effect of varying $\rho$.}
\label{fig:rho}
\end{subfigure}
\begin{subfigure}{0.48\linewidth}
\includegraphics[width=1.0\linewidth]{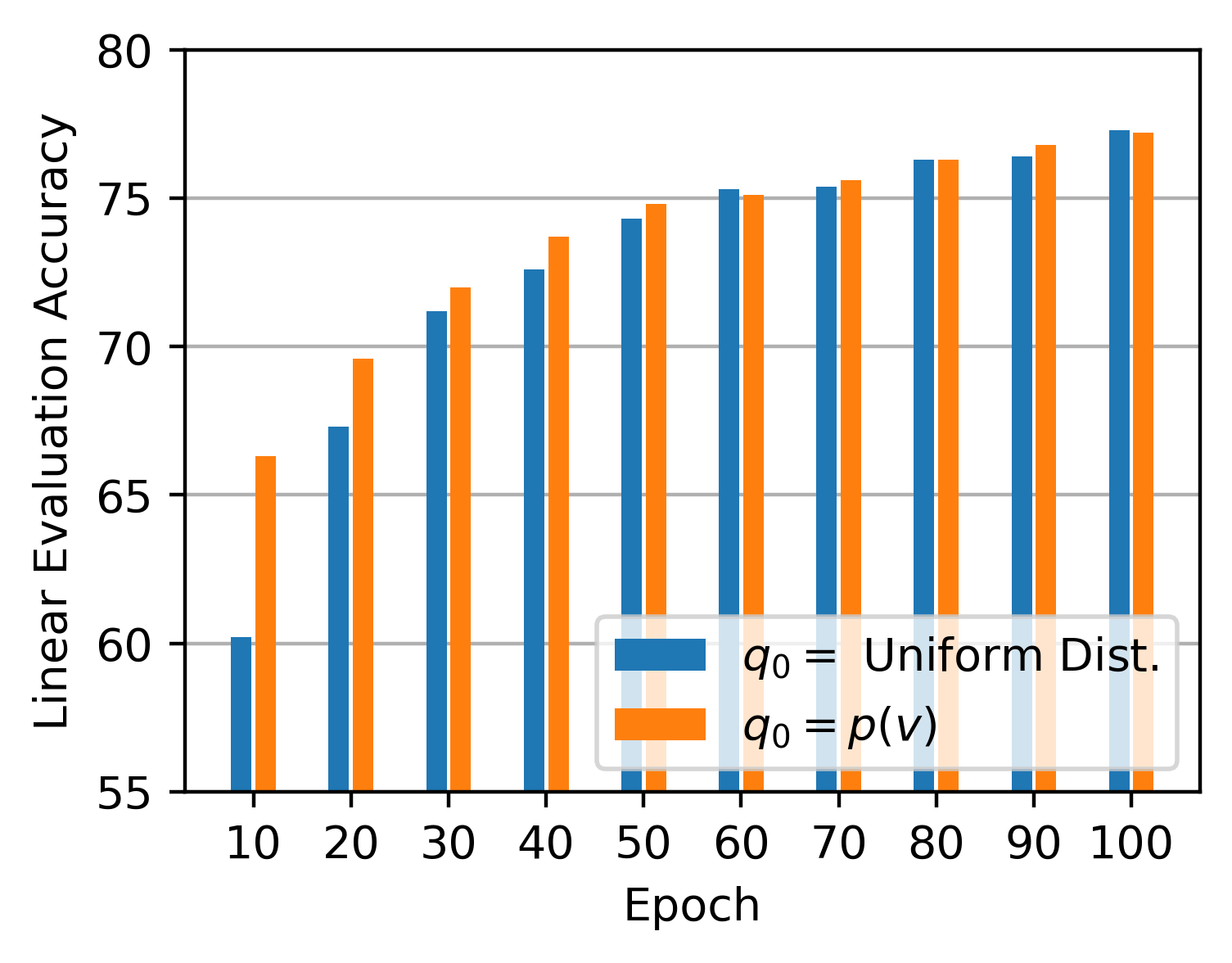}
\caption{Effect of varying $q_0$.}
\label{fig:init}
\end{subfigure}
\begin{subfigure}{\linewidth}
\includegraphics[width=1.0\linewidth]{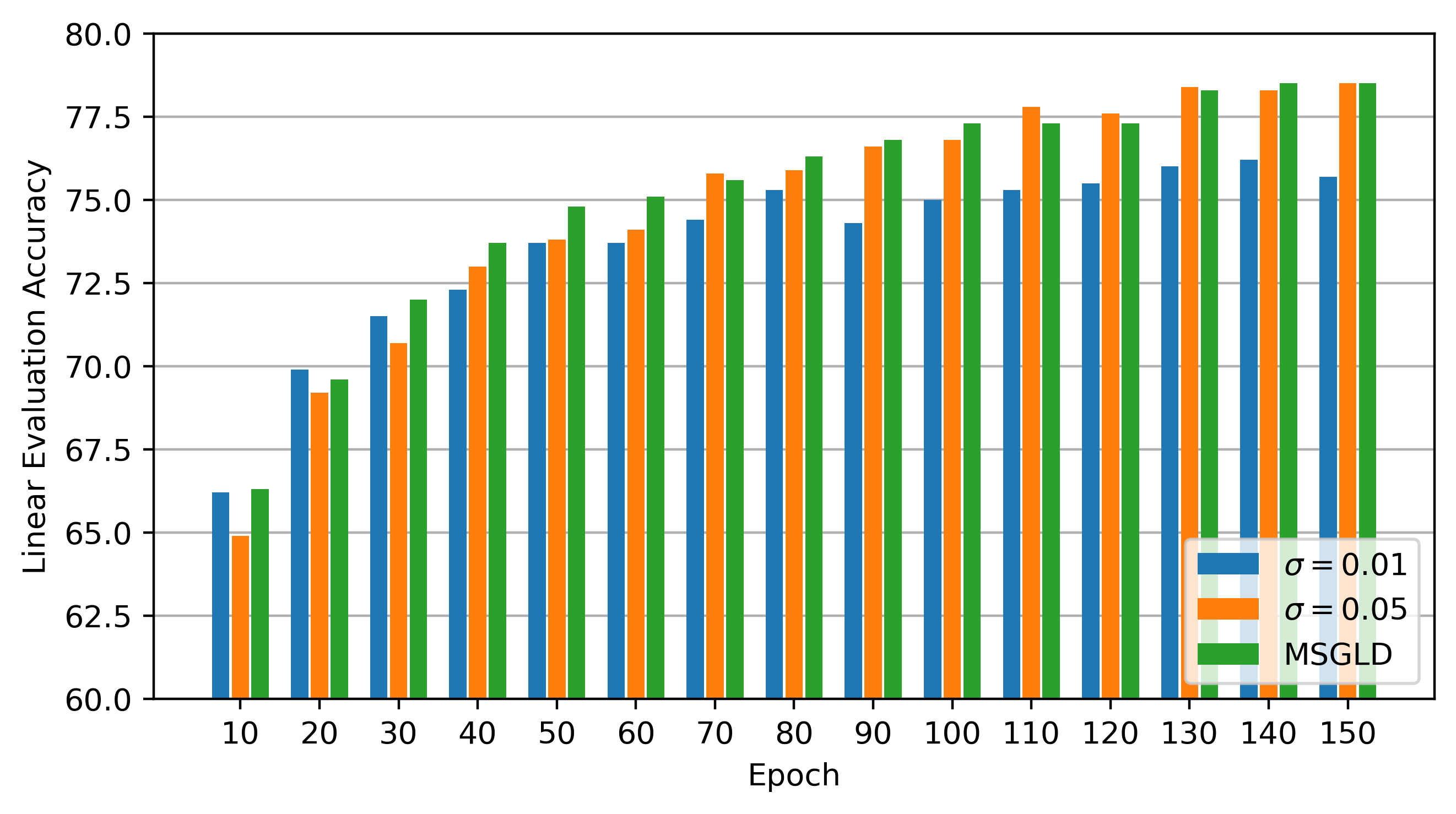}
\caption{SGLD with $\sigma \in \{0.01, 0.05\}$ and MSGLD.}
\label{fig:msgld}
\end{subfigure}
\caption{Ablation study of SGLD modifications on CIFAR10.}
\label{fig:sgld_mod}
\end{wrapfigure}

We now study the roles of the three SGLD modifications proposed in Section \ref{sec:sgld_mod}. Figure \ref{fig:sgld_mod} shows the results of varying one parameter of MSGLD while keeping the others fixed.

\textbf{Effect of reintialization frequency $\rho$.} Figure \ref{fig:rho} displays linear evaluation results for $\rho \in \{0.0, 0.2, 1.0\}$. We note that setting $\rho = 1.0$ is equivalent to removing the replay buffer. Also, setting $\rho = 0.0$ is equivalent to never reinitializing SGLD chains.

Initially, $\rho = 0.0$ shows the best performance, as SGLD quickly reaches samples of lower energy. However, learning then slows down because of the lack of diversity of samples in the replay buffer $\BB$. This implies that it is necessary to set $\rho > 0$ in order to learn good representations.

On the other hand, $\rho = 1.0$ shows slow convergence in the beginning because samples in the replay buffer are not given enough iterations to reach low energy. Although it does beat $\rho = 0.0$ at latter epochs, it still often performs worse than $\rho = 0.2$. Moreover, it is not sample-efficient compared to $\rho = 0.2$ since we have to provide an entire batch of new samples for reinitializing SGLD chains at each iteration.

Given the above observations, it is clear why the intermediate value $0.2$ is the best choice out of $\rho \in \{0.0, 0.2, 1.0\}$. $\rho = 0.2$ allows enough time for samples in the replay buffer to reach low energy while still maintaining the diversity of samples in $\BB$. Also, it is sample-efficient compared to $\rho = 1.0$.

\textbf{Effect of proposal distribution $q_0$.} Figure \ref{fig:init} compares linear evaluation accuracies with $q_0$ as the uniform distribution and $q_0 = p(v)$. We observe prominent acceleration in the initial epochs for $q_0 = p(v)$. Hence, we can conclude that this choice of proposal distribution is crucial for the high efficiency of EBCLR compared to the baseline methods in Tables \ref{table:baseline} and \ref{table:batch}.

We believe this acceleration effect can be explained by the work of Hinton \cite{hinton2002}. Specifically, let us observe that the EBM update equation \eqref{eq:ebm_grad} pushes up the energy on the model distribution $q_\theta$. In the implementation of EBCLR with $q_0 = p(v)$, however, $q_\theta$ is replaced by the distribution of samples created by a finite number of (noisy) gradient steps on real data points (see Section \ref{sec:sgld_mod}). Hence, the modified EMB update equation contains the curvature information of the data manifold. This curvature information may expedite the training process of EBCLR. For a detailed discussion on this, we refer the readers to Section 3 of the work by Hinton \cite{hinton2002}.

\textbf{Comparison of SGLD and MSGLD.} Figure \ref{fig:msgld} shows results with SGLD with $\sigma \in \{0.01, 0.05\}$ and MSGLD with $\sigma_{\min} = 0.01$ and $\sigma_{\max} = 0.05$. We note that setting $\sigma_{\min} = \sigma_{\max}$ reduces MSGLD to SGLD. We observe $\sigma = 0.01$ initially shows fast convergence but then saturates due to the lack of diversity of generated samples. On the other hand, $\sigma = 0.05$ initially has the worst performance but eventually beats $\sigma = 0.01$ since $\sigma = 0.05$ quickly explores the modes of $q_\theta$. MSGLD inherits the best of both settings. Specifically, MSGLD is as fast as $\sigma = 0.01$ in the beginning, and it does not suffer from the saturation problem.

\section{Limitations and Societal Impacts} \label{sec:limits}

\textbf{Limitations.} The main limitation of our work is of scale. While EBCLR demonstrates superior sample efficiency, it requires inner SGLD iterations (which cannot be parallelized) and a replay buffer $\BB$. These two components increase the computational burden of EBCLR. So, we found it difficult to apply EBCLR to large-scale data such as ImageNet. However, we note that inner SGLD iterations and the replay buffer are not particular limitations of EBCLR, but limitations of EBMs in general. Given the increasing efforts to overcome these limitations such as Proximal-YOPO-SGLD (for more discussion, see Appendix \ref{append:scalable}), we believe EBCLR will eventually be applicable to larger data.

\textbf{Social Impacts.} We generally expect positive outcomes from this research. Further development of EBCLR can mitigate the need for large amount of data and large batch sizes to learn good representations and ultimately lead to a reduction in resource consumption.

\section{Conclusion} \label{sec:conclusion}

In this work, we proposed EBCLR which combines contrastive learning with EBMs. This amalgamation of ideas has led to both theoretical and practical contributions. Theoretically, EBCLR associates distance in the projection space with the density of positive samples. Since the distribution of positive samples reflects the semantic similarity of images, EBCLR is capable of learning good visual representations. Practically, EBCLR is several times more sample-efficient than conventional contrastive and non-contrastive learning approaches and is robust to small numbers of negative pairs. Hence, EBCLR is applicable even in scenarios with limited data or devices. We believe that EBCLR makes representation learning available to a wider range of machine learning practitioners.

\begin{ack}
This work was supported by the National Research Foundation of Korea under Grant
NRF-2020R1A2B5B03001980, KAIST Key Research Institute (Interdisciplinary Research Group) Project, and Field-oriented Technology Development Project for Customs Administration through National Research Foundation of Korea(NRF) funded by the Ministry of Science \& ICT and Korea Customs Service(**NRF-2021M3I1A1097938**).
\end{ack}

\small

\bibliographystyle{unsrt}

\newpage

\appendix

\section{Additional Background} \label{append:background}

\subsection{Contrastive Divergence and Contrastive Learning}

Contrastive divergence, initially proposed by Hinton et al. \citep{hinton2002}, is a particular method for solving the maximum-likelihood estimation problem for EBMs. Intuitively, it proceeds by minimizing the energy function on data and maximizing the energy function on (possibly short-run) MCMC samples of the current EBM distribution through gradient ascent. In fact, we use an instance of contrastive divergence to optimize the generative term in our EBCLR objective \eqref{eq:ebclr}.

We also disambiguate contrastive learning from contrastive divergence. They are similar in the sense that they contrast certain pairs of data. Contrastive learning contrasts positive pairs and negative pairs. Contrastive divergence contrasts data and MCMC samples. However, they differ in the aspect that contrastive learning is inherently a discriminative learning method (in contrastive learning, the model “predicts” the positive pair among negative pairs) and contrastive divergence is a generative learning method.

In the context of unsupervised visual representation learning, there are several works which connect contrastive learning with generative learning \citep{wang2020,zimmermann2021,chen2021b}. For instance, Chen et al. \citep{chen2021b} combine contrastive learning with GANs to improve person re-identification. Zimmermann et al. \citep{zimmermann2021} and Wang et al. \citep{wang2020} study the distributional properties of embeddings found by contrastive learning. Our work, for the first time, explores the synergy between contrastive learning (the discriminative term in \eqref{eq:ebclr}) and EBMs / contrastive divergence (the generative term in \eqref{eq:ebclr}) for learning visual representations.


\subsection{Relation to Contrastive Energy-Based Models (CEMs) \citep{wang2022}}

According to the terms introduced in the work of Wang et al. \citep{wang2022}, our EBCLR model distribution \eqref{eq:joint} with $\tau = 1$ is identical to the model distribution of Non-Parametric CEM ((4) in \citep{wang2022}). This is because the squared $\ell_2$ norm between two unit-norm vectors $z$ and $z'$ is equivalent to the inner product between $z$ and $z'$ up to an additive constant. However, our work differs from the work of Wang et al. \citep{wang2022} in three aspects.

\textbf{Objective function.} Wang et al. \citep{wang2022} optimizes the log-likelihood of the joint model distribution $q_\theta(v,v’)$ directly whereas we use Bayes’ rule to decompose the objective into two terms, the discriminative term and the generative term. This allows us to reweight the generative term to learn better visual representations (as explored in Section \ref{sec:exp_lmda}).

\textbf{Optimization method.} Wang et al. \citep{wang2022} maximizes the log-likelihood of the joint model distribution directly via adversarial training. On the other hand, we use gradient ascent to optimize the discriminative term and contrastive divergence to optimize the generative term. Computation costs are similar, as adversarial training also requires inner iterations to generate adversarial examples at every model update step.
 
\textbf{Application of interest.} Wang et al. \citep{wang2022} focuses on enhancing image generation, while we focus on learning visual representations useful for downstream tasks.

Wang et al. \citep{wang2022} also establishes a connection between contrastive divergence and InfoNCE (Section 5.1 in \citep{wang2022}). Specifically, Wang et al. \citep{wang2022} show that when we approximate the model marginal $q_\theta(v)$ with the data distribution $p(v)$, MLE gradient for $q_\theta(v,v’)$ becomes the InfoNCE loss gradient. Surprisingly, we find this holds in our framework as well. If we set $q_\theta(v) = p(v)$ in \eqref{eq:ebclr}, the generative term in \eqref{eq:ebclr} becomes independent of $\theta$, so we end up optimizing only the discriminative term. We have shown that the discriminative term can be approximated by a contrastive learning objective, so we recover the result of Wang et al. \citep{wang2022}. In this sense, the derivations in Section 3 of our paper generalizes the connection discovered in the work of Wang et al. \citep{wang2022}, as we work with weaker assumptions (we do not set $q_\theta(v) = p(v)$ in \eqref{eq:ebclr}).

\subsection{EBMs for Learning Useful Representations}

Ranzato et al. \citep{ranzato2007} proposes to train an EBM with an autoencoder structure by minimizing the sum of an encoding loss and a decoding loss to learn an energy surface. A byproduct is that the EBM learns a compressed representation of the data, which can be used for, e.g., denoising. Our work is similar in the aspect that we use a particular EBM architecture (a triplet network) and training the EBM with our proposed method results in a compressed representation of the data which is useful for downstream tasks, e.g., classification and transfer learning. In that sense, our work is also roughly related to Denoising Autoencoders (DAEs) \citep{vincent2011} and Restricted Boltzmann Machines (RBMs) as well. Both are EBMs whose goal is to extract useful representations from data. However, we cannot say Ranzato et al. \citep{ranzato2007}, RBMs, or DAEs precede our work in relating contrastive learning to EBMs, because they do not mention anything about contrasting positive pairs against negative pairs.

\subsection{Sample Efficiency of Generative Models}

We motivate the sample efficiency of EBCLR through the lens of discriminative and generative learning. We note that standard contrastive learning (e.g., SimCLR) optimizes a discriminative objective (the model must predict positive pairs) and EBCLR optimizes a generative objective (given a pair of images, the model must output how likely it is to be a positive pair). So, we can compare the performance of discriminative models and generative models.

There is a long line of works which show (theoretically and / or empirically) that generative models can outperform discriminative ones. For instance, Efron \citep{efron1975} has theoretically shown that Normal Discriminant Analysis can be more efficient than logistic regression. Ng et al. \citep{ng2001} has also theoretically shown that in the small sample regime, na\"{i}ve Bayes can be more resistant to overfitting than logistic regression (i.e., find better solutions when given a small number of training samples).

There are also works \citep{lasserre2006,larochelle2008,grathwohl2020} that consider optimizing a weighted combination of a discriminative loss and a generative loss, just like our EBCLR objective \eqref{eq:ebclr}. They find that with an appropriate weight, the model can outperform discriminative models. In particular, Larochelle et al. \citep{larochelle2008} finds this hybrid approach to be beneficial in the limited training data setting. Grathwohl et al. \citep{grathwohl2020} uses this hybrid approach to train a classifier and find that the classifier exhibits excellent calibration, unlike other calibration methods that require additional training data.

We can also address the efficiency of generative models from the perspective of generative pre-training. Bengio et al. \citep{bengio2006} finds that pre-training a network with a generative loss brings the network weights towards a good local minimum, thus bringing better generalization. Erhan et al. \citep{erhan2010} finds that generative pre-training acts as a regularizer and that this regularization effect can be beneficial when we have a small training set.

Another explanation specific to EBCLR is that the generative loss has an effect orthogonal to the effect of the discriminative loss. To minimize the generative loss, the model needs to place low energy on data and high energy elsewhere. This can be achieved only when the model has features that accurately capture the manifold of the data. When we have a small training set, the model may easily solve the discriminative objective and overfit. By providing an auxiliary generative task, the overfitting can be mitigated. An analogous effect with Joint Energy-Based Models is demonstrated by Zhao et al. \citep{zhao2020}.

\section{Psuedocodes} \label{append:psuedocodes}

\begin{algorithm}[h]
\caption{MSGLD}
\label{alg:msgld}
\begin{algorithmic}[1]
\State \textbf{Input:} Energy $E_\theta$, initialization point $\tilde{v}$, count $\kappa_{\tilde{v}}$, step size $\alpha$, number of iterations $T$, proximity parameter $\delta$, decay parameter $K$, variance bounds $\sigma_{\min}^2$, $\sigma_{\max}^2$
\State $\sigma = \sigma_{\min} + (\sigma_{\max} - \sigma_{\min}) \cdot [1 - \kappa_{\tilde{v}} / K]_+$
\For {$t = 0, 1, 2, \ldots T - 1$}
\State Sample $\epsilon \sim \NN(0,\sigma^2)$
\State $\tilde{v} \leftarrow \tilde{v} - \alpha \cdot \clamp\{\nabla_{v} E_\theta(\tilde{v}),\delta\} + \epsilon$
\EndFor
\State $\kappa_{\tilde{v}} \leftarrow \kappa_{\tilde{v}} + 1$
\end{algorithmic}
\end{algorithm}

\begin{algorithm}[h]
\caption{EBCLR}
\label{alg:ebclr}
\begin{algorithmic}[1]
\State \textbf{Input:} DNN $f_\theta$, batch size $N$, generative term weight $\lambda$, replay buffer $\BB$, reinit. frequency $\rho$
\While {not converged}
\State Sample $\{(v_n,v_n')\}_{n = 1}^N$ from $p(v,v')$
\State Calculate the disc. term with Eq. \eqref{eq:q_bar_approx}
\State Calculate gradient of the disc. term
\State Sample $\{\tilde{v}_n\}_{n = 1}^N$ from $\BB$ with prob. $1 - \rho$ and from $p(v)$ with prob. $\rho$
\State Calculate $E_\theta(v;\{v_m'\}_{m = 1}^N)$ with Eq. \eqref{eq:energy_approx}
\State Update $\{\tilde{v}_n\}_{n = 1}^N$ with MSGLD in Algo. \ref{alg:msgld}
\State Calculate gradient of the gen. term with Eq. \eqref{eq:ebm_grad}
\State Update $\theta$ via gradient ascent
\State Update $\BB$ with MSGLD samples $\{\tilde{v}_n\}_{n = 1}^N$
\EndWhile
\end{algorithmic}
\end{algorithm}

\newpage

\section{Missing Proofs} \label{append:proofs}

\subsection{Proof of Theorem \ref{thm:2}} \label{append:thm2_proof}

\begin{proof}
We first observe that
\begin{align*}
\frac{1}{Z(\theta)} \exp\{-E_\theta(v)\} &= \frac{1}{Z(\theta)} \int e^{-\|z - z'\|^2 / \tau} \, dv' \\
&= \int \frac{1}{Z(\theta)} e^{-\|z - z'\|^2 / \tau} \, dv' \\
&= \int q_\theta(v,v') \, dv' \\
&= q_\theta(v)
\end{align*}
which proves the relation \eqref{eq:marginal_ebm}. We also have
\begin{align*}
\log q_\theta(v) = -\log Z(\theta) - E_\theta(v)
\end{align*}
and so
\begin{align*}
\nabla_\theta \log q_\theta(v) &= - \frac{1}{Z(\theta)} \nabla_\theta Z(\theta) - \nabla_\theta E_\theta(v) \\
&= - \frac{1}{Z(\theta)} \nabla_\theta \int \exp\{-E_\theta(v)\} \, dv - \nabla_\theta E_\theta(v) \\
&= - \frac{1}{Z(\theta)} \int \nabla_\theta \exp\{-E_\theta(v)\} \, dv - \nabla_\theta E_\theta(v) \\
&= - \frac{1}{Z(\theta)} \int \{- \nabla_\theta E_\theta(v)\} \cdot \exp\{-E_\theta(v)\} \, dv - \nabla_\theta E_\theta(v) \\
&= \int \{\nabla_\theta E_\theta(v)\} \cdot \frac{1}{Z(\theta)} \exp\{-E_\theta(v)\} \, dv - \nabla_\theta E_\theta(v) \\
&= \int \{\nabla_\theta E_\theta(v)\} \cdot q_\theta(v) \, dv - \nabla_\theta E_\theta(v) \\
&= \EE_{q_\theta}[\nabla_\theta E_\theta(v)] - \nabla_\theta E_\theta(v).
\end{align*}
Taking expectation w.r.t. $p(v)$ establishes the relation \eqref{eq:marginal_grad}.
\end{proof}

\subsection{Another Theoretical Justification for EBCLR} \label{append:another}

Here we provide another theoretical justification on how solving the EBCLR objective \eqref{eq:ebclr} with $\lambda = 1$ causes $q_\theta(v,v')$ to approximate $p(v,v')$. It is known that optimizing the first term of \eqref{eq:ebclr} with \eqref{eq:q_bar_approx} in place of $q_\theta(v'\mid v)$ will cause \eqref{eq:q_bar_approx} to approximate $p(v' \mid v)$ \citep{zhang2021}. Also, optimizing the second term of \eqref{eq:ebclr} with \eqref{eq:approx2} or \eqref{eq:approx1} in place of $q_\theta(v)$ will cause \eqref{eq:approx2} or \eqref{eq:approx1} to be proportional to $p(v)$ (indeed, in Table \ref{table:fid}, we see SGLD samples of the EBCLR marginal $q_\theta(v)$ approximated by \eqref{eq:approx1} achieve a non-trivial FID score). So, the product of \eqref{eq:approx2} or \eqref{eq:approx1} with \eqref{eq:q_bar_approx} will be proportional to $p(v,v')$. Moreover, by construction, the product of \eqref{eq:approx2} or \eqref{eq:approx1} with \eqref{eq:q_bar_approx} is (approximately) $q_\theta(v,v')$. Hence, optimizing the EBCLR objective \eqref{eq:ebclr} will cause $q_\theta(v,v')$ to model $p(v,v')$.

\section{Complete Training and Evaluation Details} \label{append:details}

\textbf{Baseline methods and datasets.} The baseline methods are SimCLR, MoCo v2, SimSiam, and BYOL. The hyper-parameters are chosen closely following the original works \cite{chen2020,chen2020moco,chen2021,grill2020}. We use four datasets: MNIST \cite{mnist}, Fashion MNIST (FMNIST) \cite{fmnist}, CIFAR10, and CIFAR100 \cite{cifar}.

\textbf{Data transformation.} For fair comparison, we use the stochastic data transformation proposed by Chen et. al \cite{chen2020} for all methods. In the case of grayscale images, we only use the random cropping augmentation. For EBCLR, we also add small Gaussian noise $\NN(0,0.03^2)$ to stabilize training \cite{du2019,grathwohl2020,yang2021}. We remark that Gaussian noise of standard deviation $0.03$ is nearly invisible to the human eye.

\textbf{DNN architecture.} We decompose $f_\theta = \pi_\theta \circ \phi_\theta$ where $\phi_\theta$ is the encoder network and $\pi_\theta$ is the projection network. Rather than directly using the output of $f_\theta$ for downstream tasks, we follow previous works \cite{chen2020,he2020,oord2018,tian2019,tian2020,grill2020,chen2021} and use the output of $\phi_\theta$ instead.

In our experiments, we set $\phi_\theta$ to be a ResNet-18 \cite{he2016} up to the global average pooling layer. The architecture of $\pi_\theta$ and $\pi_\theta$ to be a 2-layer MLP with output dimension $128$. However, we remove batch normalization because batch normalization hurts SGLD \cite{du2019}. We also replace ReLU activations with leaky ReLU to expedite the convergence of SGLD. For the baseline methods, we use the settings proposed in the original works while keeping the backbone fixed to be ResNet-18.

\textbf{SGLD settings.} The default parameters for MSGLD are $|\BB| = 50k$, $\rho = 0.2$, $T = 10$, $\alpha = 0.05$, $\delta = 1.0$, $K = 3$, $\sigma_{\min} = 0.01$, and $\sigma_{\max} = 0.05$. We use $\rho = 0.4$ for MNIST, $\rho = 0.6$ for FMNIST, and $\rho = 0.2$ for CIFAR10 and CIFAR100. At the beginning of training, we fill the buffer $\BB$ entirely with proposal distribution samples.

\textbf{Training.} The temperature is $\tau = 0.1$ for EBCLR. For EBCLR, as we cannot use batch normalization, it is difficult to apply SGD with momentum. Hence, we use the Adam optimizer \cite{kingma2015}. The learning rate is $0.0002$ if the batch size is $128$ and $0.0001$ if the batch size is smaller than $128$. We also weakly regularize the squared $\ell_2$-norm of the outputs of $f_\theta$ with weight $0.001$. For the baseline methods, we closely follow the settings of previous works \cite{chen2020,chen2020moco,wang2021}. Specifically, we use SGD with momentum 0.9 and learning rate of $0.03 \cdot (\textit{batch size} / 256)$ and weight decay $0.0001$. Each model is trained with a single GeForce RTX 3090.

\textbf{Evaluation.} We assess the quality of the representations by training a linear classifier on top of frozen $\phi_\theta$. The linear classifier is trained with Adam for $200$ epochs with batch size $512$. The learning rate $\eta$ is found by grid search for $\log_{10} \eta$ in $[-4,-1]$.

\section{Additional Experiments} \label{append:add_exp}

\subsection{Section \ref{sec:exp_comp} with Longer Training} \label{append:exp_comp}

We repeat the experiments in Section \ref{sec:exp_comp} by training the models for 200 epochs. For EBCLR, the number of SGLD iterations $T$ is increased by $5$ when SGLD is unable to generate samples with sufficiently high energy. Tables \ref{table:baseline_200} and \ref{table:transfer_200} describe the results. We again note that EBCLR consistently beats all baseline methods and demonstrates high efficiency.

\begin{table*}[h!]
\centering
\resizebox{0.8\textwidth}{!}{
\begin{tabular}{c c c c c c c c c}
\toprule
Dataset & \multicolumn{2}{c}{\textbf{MNIST}} & \multicolumn{2}{c}{\textbf{FMNIST}} & \multicolumn{2}{c}{\textbf{CIFAR10}} & \multicolumn{2}{c}{\textbf{CIFAR100}} \\
\cmidrule(lr){1-1} \cmidrule(lr){2-3} \cmidrule(lr){4-5} \cmidrule(lr){6-7} \cmidrule(lr){8-9}
Statistic & Accuracy & Rel. Eff. & Accuracy & Rel. Eff. & Accuracy & Rel. Eff. & Accuracy & Rel. Eff. \\
\cmidrule{1-9}
SimSiam & $98.0$ & $0.025$ & $87.3$ & $0.05$ & $72.8$ & $0.175$ & $42.9$ & $0.125$ \\
BYOL    & $\mathbf{99.4}$ & $0.5$ & $89.5$ & $0.25$ & $75.3$ & $0.325$ & $45.7$ & $0.2$ \\
SimCLR  & $99.1$ & $0.05$ & $88.9$ & $0.05$ & $72.7$ & $0.175$ & $41.9$ & $0.05$ \\
MoCo v2 & $98.9$ & $0.05$ & $89.4$ & $0.25$ & $67.4$ & $0.075$ & $45.0$ & $0.2$ \\
\cmidrule{1-9}
EBCLR   & $\mathbf{99.4}$ & -- & $\mathbf{90.8}$ & -- & $\mathbf{80.0}$ & -- & $\mathbf{51.6}$ & -- \\
\bottomrule
\end{tabular}}
\caption{Linear evaluation accuracy and efficiency relative to EBCLR. Efficiency of a method relative to EBCLR is calculated by the following formula: (number of epochs used by EBCLR to reach the final accuracy of the method) / (total number of training epochs).}
\label{table:baseline_200}
\end{table*}

\begin{table*}[h!]
\centering
\resizebox{0.5\textwidth}{!}{
\begin{tabular}{c c c c c}
\toprule
Direction & \textbf{M} $\rightarrow$ \textbf{FM} & \textbf{FM} $\rightarrow$ \textbf{M} & \textbf{C10} $\rightarrow$ \textbf{C100} & \textbf{C100} $\rightarrow$ \textbf{C10} \\
\cmidrule{1-5}
SimSiam & $86.5$ & $95.2$ & $43.2$ & $67.0$ \\
BYOL    & $87.1$ & $97.3$ & $47.7$ & $73.6$ \\
SimCLR  & $86.9$ & $97.2$ & $45.7$ & $65.4$ \\
MoCo v2 & $86.2$ & $97.6$ & $41.2$ & $68.7$ \\
\cmidrule{1-5}
EBCLR   & $\mathbf{87.3}$ & $\mathbf{98.5}$ & $\mathbf{48.4}$ & $\mathbf{74.2}$ \\
\bottomrule
\end{tabular}}
\caption{Comparison of transfer learning results in the linear evaluation setting. Left side of the arrow is the dataset than the encoder was pre-trained on, and right side of the arrow is the dataset that linear evaluation was performed on. We use the following abbreviations. \textbf{M} : MNIST, \textbf{FM} : FMNIST, \textbf{C10} : CIFAR10, \textbf{C100} : CIFAR100.}
\label{table:transfer_200}
\end{table*}

\newpage

\subsection{Section \ref{sec:exp_neg} with Longer Training} \label{append:exp_neg}

We repeat the experiments in Section \ref{sec:exp_neg} by training the models for 200 epochs. For EBCLR, the number of SGLD iterations $T$ is increased by $5$ when SGLD is unable to generate samples with sufficiently high energy. Table \ref{table:batch_200} describes the results. We observe similar patterns as in the case of training for 100 epochs. In the case of CIFAR10, SimCLR seems to have a pattern agnostic to the batch size. However, we note that there is a significant gap between the performance of SimCLR with batch size $256$ (see Table \ref{table:baseline_200}) and batch sizes in $\{16, 64, 128\}$.

Interestingly, for SimCLR with small batch sizes, we observe that training longer with small batch sizes leads to worse performance than training only for $100$ epochs. This is particularly prominent on CIFAR10 and CIFAR100 (compare Tables \ref{table:batch} and \ref{table:batch_200}). We speculate that this is because, with small batches, SimCLR is unable to provide hard negatives, and this leads to overfitting of the DNN. On the other hand, EBCLR does not suffer from this phenomenon. This again demonstrates the importance of the generative term.

\begin{table*}[h!]
\centering
\resizebox{0.8\textwidth}{!}{
\begin{tabular}{c c c c c c c c c c c c c}
\toprule
Dataset & \multicolumn{3}{c}{\textbf{MNIST}} & \multicolumn{3}{c}{\textbf{FMNIST}} & \multicolumn{3}{c}{\textbf{CIFAR10}} & \multicolumn{3}{c}{\textbf{CIFAR100}} \\
\cmidrule(lr){1-1} \cmidrule(lr){2-4} \cmidrule(lr){5-7} \cmidrule(lr){8-10} \cmidrule(lr){11-13}
Batch Size & $16$ & $64$ & $128$ & $16$ & $64$ & $128$ & $16$ & $64$ & $128$ & $16$ & $64$ & $128$ \\
\cmidrule{1-13}
SimCLR & $98.7$ & $98.7$ & $99.0$ & $86.9$ & $88.5$ & $88.9$ & $65.3$ & $64.0$ & $63.8$ & $33.4$ & $39.6$ & $43.0$ \\
EBCLR  & $\mathbf{99.4}$ & $\mathbf{99.4}$ & $\mathbf{99.4}$ & $\mathbf{89.8}$ & $\mathbf{90.2}$ & $\mathbf{90.2}$ & $\mathbf{79.2}$ & $\mathbf{80.2}$ & $\mathbf{80.0}$ & $\mathbf{51.0}$ & $\mathbf{52.4}$ & $\mathbf{51.6}$ \\
\cmidrule{1-13}
Rel. Eff. & $0.3$ & $0.05$ & $0.45$ & $0.05$ & $0.1$ & $0.1$ & $0.05$ & $0.05$ & $0.05$ & $0.025$ & $0.1$ & $0.15$ \\
\bottomrule
\end{tabular}}
\caption{Linear evaluation accuracies and efficiencies relative to EBCLR with various batch sizes. Efficiency of SimCLR relative to EBCLR is calculated by the following formula: (number of epochs used by EBCLR with the same batch size to reach the final accuracy of SimCLR) / (total number of training epochs).}
\label{table:batch_200}
\end{table*}

\subsection{Comparison of Approximations} \label{sec:approx_comp}

\begin{figure}[h!]
\centering
\includegraphics[width=0.55\linewidth]{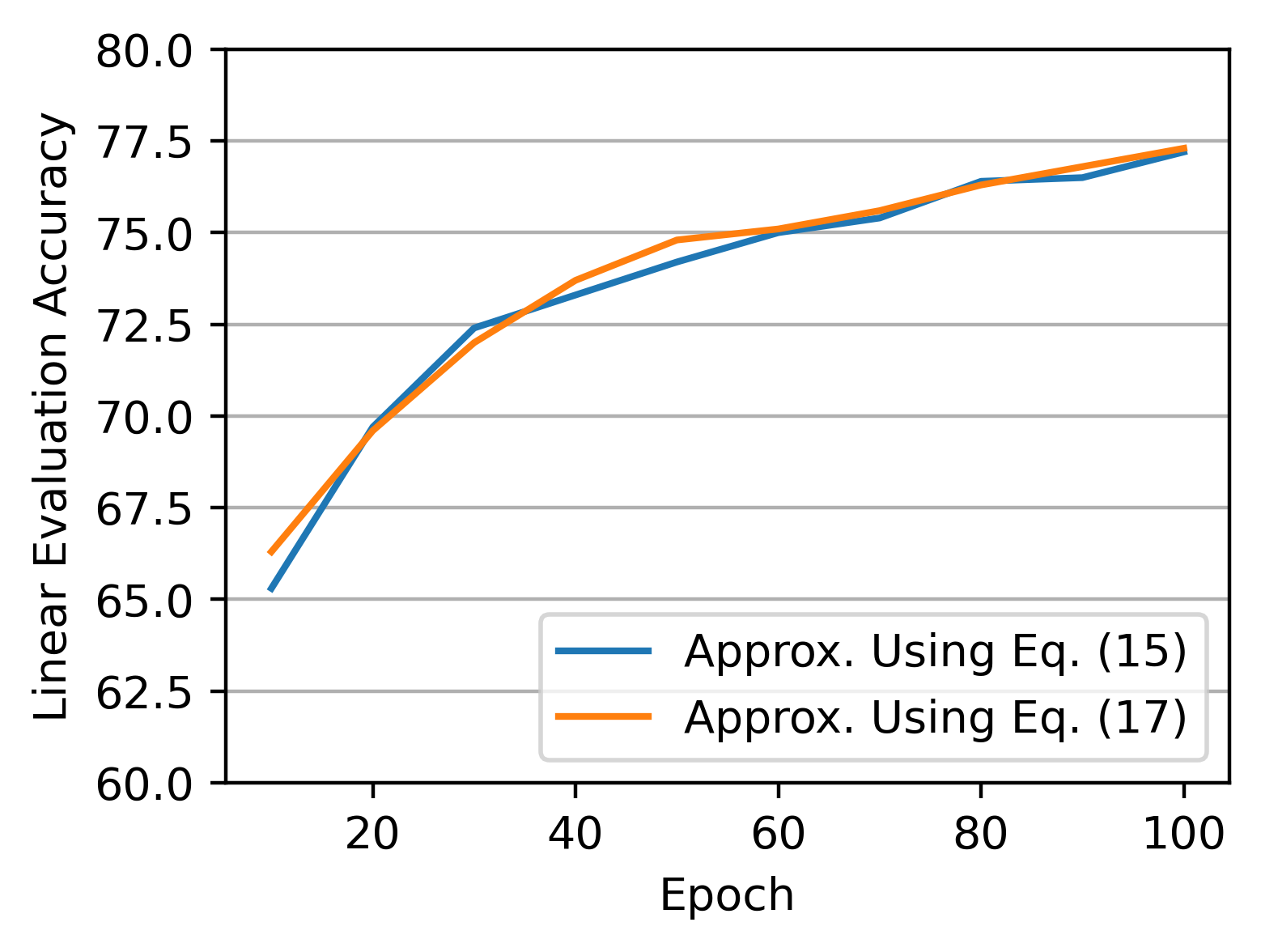}
\caption{Comparison of different approximations of the generative term, demonstrated on CIFAR10.}
\label{fig:approx}
\end{figure}

We can either use Equation \eqref{eq:approx2} or Equation \eqref{eq:approx1} when approximating the generative term. Figure \ref{fig:approx} shows results for both approximations on CIFAR10. We observe identical performance for both approximations, which implies that the generative term is robust to the number of samples used in the calculation of the empirical mean. This justifies our choice of using the simpler Equation \eqref{eq:approx1} to approximate the marginal distribution $q(v)$.

\newpage

\subsection{Evaluation with KNN Classifiers} \label{append:knn}

We also reproduce Tables \ref{table:baseline}, \ref{table:transfer}, and \ref{table:batch} using K-nearest neighbour (KNN) classifiers. We followed the evaluation protocol of Caron et al. \citep{caron2021} and used a weighted KNN classifier with $K = 20$. The results are given below. In Table \ref{table:knn_baseline}, except for the case of BYOL on MNIST, we see that EBCLR beats all baselines again by a non-trivial margin. EBCLR is also shown to be sample efficient. In Table \ref{table:knn_transfer}, we see EBCLR outperforms all baselines on the task of transfer learning. In Table \ref{table:knn_batch}, we see EBCLR is robust to small batch sizes. This leads to the lower efficiency of SimCLR compared to EBCLR. On CIFAR10 and CIFAR100, relative efficiency does not decrease below 0.05 since we saved checkpoints every 5 epochs. We can reasonably expect SimCLR to have lower relative efficiency at smaller batch sizes if we had saved checkpoints every epoch.

\begin{table*}[h!]
\centering
\resizebox{0.9\textwidth}{!}{
\begin{tabular}{c c c c c c c c c}
\toprule
Dataset & \multicolumn{2}{c}{\textbf{MNIST}} & \multicolumn{2}{c}{\textbf{FMNIST}} & \multicolumn{2}{c}{\textbf{CIFAR10}} & \multicolumn{2}{c}{\textbf{CIFAR100}} \\
\cmidrule(lr){1-1} \cmidrule(lr){2-3} \cmidrule(lr){4-5} \cmidrule(lr){6-7} \cmidrule(lr){8-9}
Statistic & Accuracy & Rel. Eff. & Accuracy & Rel. Eff. & Accuracy & Rel. Eff. & Accuracy & Rel. Eff. \\
\cmidrule{1-9}
SimSiam & $94.5$ & $0.05$ & $80.4$ & $0.01$ & $59.6$ & $0.1$ & $23.9$ & $0.05$ \\
BYOL & $\mathbf{98.8}$ & -- & $84.4$ & $0.2$ & $62.3$ & $0.15$ & $31.3$ & $0.1$ \\
SimCLR  & $97.4$ & $0.15$ & $84.1$ & $0.15$ & $57.2$ & $0.05$ & $29.5$ & $0.1$ \\
MoCo v2 & $94.5$ & $0.05$ & $83.1$ & $0.1$ & $54.1$ & $0.05$ & $26.0$ & $0.05$ \\
\cmidrule{1-9}
EBCLR   & $98.1$ & -- & $\mathbf{86.6}$ & -- & $\mathbf{71.4}$ & -- & $\mathbf{39.4}$ & -- \\
\bottomrule
\end{tabular}}
\caption{KNN accuracy and efficiency relative to EBCLR. Efficiency of a method relative to EBCLR is calculated by the following formula: (number of epochs used by EBCLR to reach the final accuracy of the method) / (total number of training epochs).}
\label{table:knn_baseline}
\end{table*}

\begin{table*}[h!]
\centering
\resizebox{0.6\textwidth}{!}{
\begin{tabular}{c c c c c}
\toprule
Direction & \textbf{M} $\rightarrow$ \textbf{FM} & \textbf{FM} $\rightarrow$ \textbf{M} & \textbf{C10} $\rightarrow$ \textbf{C100} & \textbf{C100} $\rightarrow$ \textbf{C10} \\
\cmidrule{1-5}
SimSiam & $80.3$ & $85.1$ & $26.1$ & $52.6$ \\
BYOL    & $82.6$ & $92.4$ & $29.3$ & $63.7$ \\
SimCLR  & $80.7$ & $87.6$ & $25.4$ & $56.0$ \\
MoCo v2 & $78.7$ & $89.7$ & $23.8$ & $52.2$ \\
\cmidrule{1-5}
EBCLR   & $\mathbf{83.4}$ & $\mathbf{95.5}$ & $\mathbf{35.7}$ & $\mathbf{65.4}$ \\
\bottomrule
\end{tabular}}
\vspace{1em}
\caption{Comparison of transfer learning results in the KNN evaluation setting. Left side of the arrow is the dataset than the encoder was pre-trained on, and right side of the arrow is the dataset that KNN evaluation was performed on. We use the following abbreviations. \textbf{M} : MNIST, \textbf{FM} : FMNIST, \textbf{C10} : CIFAR10, \textbf{C100} : CIFAR100.}
\label{table:knn_transfer}
\end{table*}

\begin{table*}[h!]
\centering
\resizebox{0.9\textwidth}{!}{
\begin{tabular}{c c c c c c c c c c c c c}
\toprule
Dataset & \multicolumn{3}{c}{\textbf{MNIST}} & \multicolumn{3}{c}{\textbf{FMNIST}} & \multicolumn{3}{c}{\textbf{CIFAR10}} & \multicolumn{3}{c}{\textbf{CIFAR100}} \\
\cmidrule(lr){1-1} \cmidrule(lr){2-4} \cmidrule(lr){5-7} \cmidrule(lr){8-10} \cmidrule(lr){11-13}
Batch Size & $16$ & $64$ & $128$ & $16$ & $64$ & $128$ & $16$ & $64$ & $128$ & $16$ & $64$ & $128$ \\
\cmidrule{1-13}
SimCLR & $97.1$ & $97.7$ & $97.4$ & $82.3$ & $83.2$ & $83.3$ & $52.8$ & $56.0$ & $56.8$ & $22.1$ & $22.7$ & $28.8$ \\
EBCLR & $\mathbf{98.5}$ & $\mathbf{98.3}$ & $\mathbf{98.1}$ & $\mathbf{85.0}$ & $\mathbf{85.9}$ & $\mathbf{86.6}$ & $\mathbf{72.2}$ & $\mathbf{72.6}$ & $\mathbf{71.4}$ & $\mathbf{40.6}$ & $\mathbf{41.4}$ & $\mathbf{39.4}$ \\
\cmidrule{1-13}
Rel. Eff. & $0.05$ & $0.05$ & $0.15$ & $0.05$ & $0.15$ & $0.1$ & $0.05$ & $0.05$ & $0.05$ & $0.05$ & $0.05$ & $0.05$ \\
\bottomrule
\end{tabular}}
\caption{KNN accuracies and efficiencies relative to EBCLR with various batch sizes. Efficiency of SimCLR relative to EBCLR is calculated by the following formula: (number of epochs used by EBCLR with the same batch size to reach the final accuracy of SimCLR) / (total number of training epochs).}
\label{table:knn_batch}
\end{table*}

\newpage

\subsection{Generative Performance of EBCLR} \label{append:exp_gen}

In this section, we explore the effect of the discriminative loss on the generative performance of EBCLR. To this end, we solved
\begin{align}
\max_\theta \gamma \EE_p [\log q_\theta(v' \mid v)] + \EE_p[\log q_\theta(v)]
\end{align}
for various values of $\gamma$ on FMNIST. Recall that in the original EBCLR objective, weight $\lambda$ comes in front of the generative term. We then measured the Fr\'{e}chet Inception Distance (FID, lower is better) between samples from the true marginal $p(v)$ and SGLD samples from the EBM marginal $q_\theta(v)$ approximated by \eqref{eq:q_approx} / \eqref{eq:approx1}. The results are shown in Table \ref{table:fid}.

Interestingly, we found using an appropriate $\gamma > 0$ led to improvements in FID over $\gamma = 0$. In other words, contrastive learning did help improve the generative performance of the model. However, using an excessively large $\gamma$ led to a deterioration of the performance. This is analogous to the result of Section \ref{sec:exp_lmda} where we observed using an appropriate weight on the generative term led to improvements over using only the discriminative term (contrastive loss). The difference is that to achieve optimal generative performance, the generative term weight needs to be larger (i.e., $\gamma < 1$) whereas to achieve optimal discriminative performance, the discriminative term weight needs to be larger (i.e., $\lambda < 1$ in \eqref{eq:ebclr}). We speculate that using a larger model could mitigate this trade-off, but this topic is beyond the scope of this work.

\begin{table*}[h!]
\centering
\resizebox{0.45\textwidth}{!}{
\begin{tabular}{c c c c c c}
\toprule
$\gamma$ & 0 & 0.01 & 0.1 & 1 & 10 \\
\cmidrule{1-6}
FID $\downarrow$ & $40.78$ & $9.68$ & $7.68$ & $17.03$ & $139.19$ \\
\bottomrule
\end{tabular}}
\caption{Generative performance of EBCLR for various values of $\gamma$.}
\label{table:fid}
\end{table*}

\section{A Discussion on Making EBCLR Scalable} \label{append:scalable}

We could consider three alternative losses / methods for training EBCLR: noise contrastive estimation (NCE) \citep{gutmann2010}, VERA \citep{grathwohl2021}, and score matching \citep{hyvarinen2005}. Although the NCE loss does not depend on MCMC, it requires specification of a noise distribution whose normalized density can be easily evaluated. In most cases, such noise distribution samples are easily distinguishable from natural image samples and lead to the density chasm problem \citep{rhodes2020}, making optimization difficult. Telescoping density-ratio estimation \citep{rhodes2020} attempts to mitigate this problem, but it does not seem to scale well to large data. Flow contrastive estimation \citep{gao2020} uses a flow model to define the noise distribution and trains a flow model along with the EBM. Though flow contrastive estimation requires an auxiliary flow model, we believe this is the most promising alternative to MCMC. In a similar sense, VERA, which uses a generator to sample from the EBM, seems promising as well. Yet, flow contrastive estimation and VERA also did not demonstrate scalability to large-scale datasets such as ImageNet. Score matching matches the gradient of model log-density to the gradient of data log-density, so it does not rely on MCMC. However, it requires calculation of the Hessian, so in the deep learning setting, this approach is also quite expensive.

\end{document}